\documentclass[10pt,twocolumn,letterpaper]{article}

\usepackage[pagenumbers]{cvpr}

\usepackage{graphicx}
\usepackage{amsmath}
\usepackage{amssymb}
\usepackage{booktabs}

\usepackage[pagebackref,breaklinks,colorlinks]{hyperref}

\usepackage[capitalize]{cleveref}
\crefname{section}{Sec.}{Secs.}
\Crefname{section}{Section}{Sections}
\Crefname{table}{Table}{Tables}
\crefname{table}{Tab.}{Tabs.}

\def\cvprPaperID{7} 
\def\confName{CVPR}
\def\confYear{2022}

\begin{document}

\title{Blind Non-Uniform Motion Deblurring using Atrous Spatial Pyramid Deformable Convolution and Deblurring-Reblurring Consistency}

\author{Dong Huo, Abbas Masoumzadeh, Yee-Hong Yang\\
Department of Computing Science\\
University of Alberta, Edmonton, Canada\\
{\tt\small \{dhuo, a.masoumzadeh, herberty\}@ualberta.ca}
}
\maketitle

\begin{abstract}
   Many deep learning based methods are designed to remove non-uniform (spatially variant) motion blur caused by object motion and camera shake without knowing the blur kernel. Some methods directly output the latent sharp image in one stage, while others utilize a multi-stage strategy (\eg multi-scale, multi-patch, or multi-temporal) to gradually restore the sharp image. However, these methods have the following two main issues: 1) The computational cost of multi-stage is high; 2) The same convolution kernel is applied in different regions, which is not an ideal choice for non-uniform blur. Hence, non-uniform motion deblurring is still a challenging and open problem. In this paper, we propose a new architecture which consists of multiple Atrous Spatial Pyramid Deformable Convolution (ASPDC) modules to deblur an image end-to-end with more flexibility. Multiple ASPDC modules implicitly learn the pixel-specific motion with different dilation rates in the same layer to handle movements of different magnitude. To improve the training, we also propose a reblurring network to map the deblurred output back to the blurred input, which constrains the solution space. Our experimental results show that the proposed method outperforms state-of-the-art methods on the benchmark datasets. The code is available at \href{https://github.com/Dong-Huo/ASPDC}{https://github.com/Dong-Huo/ASPDC}.
\end{abstract}

\section{Introduction}
When we are taking photos using a camera, especially the one on a mobile device, non-uniform motion blur is one of the most common types of undesirable artifacts caused by object motion and camera shake~\cite{gong2017motion}. Removing such blur to recover the original sharp image plays a critical role in many high-level vision tasks, e.g. computational photography~\cite{lukac2017computational}, image classification~\cite{rawat2017deep}, object detection~\cite{zhao2019object}, and face recognition~\cite{adjabi2020past}, because motion blur severely degrades the image quality. Both Zhang \etal~\cite{zhang2020deblurring} and Nah \etal~\cite{nah2017deep} claim that motion blur can be regarded as the temporal integration of multiple sharp snapshots during the exposure time, and can be formulated as:
\begin{gather} 
I_b= g\left(\frac{1}{T}\int_{t=0}^{T}I_{S(t)}dt\right),
\label{eqn:integration_model}
\end{gather}  
in which $I_b$ is the blurred image of the dynamic scene, $T$ is the period of the exposure time, $I_{S(t)}$ is the sharp snapshot at timestamp $t$ and $g()$ represents the Camera Response Function (CRF). 

In addition to the above model, some works model the non-uniform motion deblurring as a linear transformation. Indeed, Bahat \etal~\cite{bahat2017non} formulate the problem using the following equation:
\begin{gather}
I_b= I_s \circledast k + n,
\label{eqn:convolution_model}
\end{gather}  
where $I_b$ and $I_s$ represent the blurred and sharp image, respectively, $n$ is the additive noise, $k$ is the spatially variant blur kernel matrix which is different from some of the deblurring methods using a uniform blur kernel~\cite{dong2018learning, kaufman2020deblurring, ren2020neural}. Each column of $k$ represents a blur kernel for the corresponding pixel in $I_s$. The blur kernel matrix $k$ is applied to the sharp image by the matrix multiplication operator $\circledast$. The objective is to find the $I_s$ given $I_b$, assuming that $k$ and $n$ are unknown. 

To handle the blind deblurring problem, many conventional methods attempt to first estimate the blur kernel, then use it to recover the sharp image with some hand-crafted priors of $I_s$ and $k$~\cite{gupta2010single, liu2014blind, pan2016l_0, pan2016blind, strong2003edge, yan2017image,  zuo2016learning}. However, Ren \etal~\cite{ren2020neural} claim that hand-crafted priors are insufficient to recover the ideal sharp image, and an improper prior can even lead to an incorrect kernel. Recently, deep learning based methods~\cite{aljadaany2019douglas, kupyn2018deblurgan, kupyn2019deblurgan, nah2017deep, park2019multi, purohit2020region, suin2020spatially, tao2018scale, yuan2020efficient, zhang2019deep, zhang2018dynamic} significantly improve the performance of non-uniform motion deblurring. The priors of $I_s$ and $k$ are implicitly learned by the network, which outputs the deblurred result directly, and bypasses the need to estimate the blur kernel. However, existing deep learning methods suffer from two main issues: 1) Many state-of-the-art (SOTA) methods utilize a multi-stage strategy, such as multi-scale~\cite{nah2017deep, tao2018scale}, multi-patch~\cite{suin2020spatially, zhang2019deep}, or multi-temporal~\cite{park2019multi}, all of which increase the computational cost. 2) The same convolution filters~\cite{kupyn2019deblurgan} or filters of the same receptive fields~\cite{purohit2020region} are applied to different regions of an image. To overcome such issue, extremely deep and wide networks have been exploited to improve the generalization on different levels of blur.

To address the above issues, we propose a new \textit{Atrous Spatial Pyramid Deformable Convolution} (ASPDC) module for region-specific convolution and for integrating features from different sizes of receptive fields, which is more suitable for non-uniform deblurring. We also propose a new reblurring network to reblur the output, which is helpful in constraining the solution space~\cite{guo2020closed} of deblurring with a new deblurring-reblurring consistency loss. Note that our reblurring network is used during training only, which needs both of the blurred image and the corresponding sharp (deblurred) image. More details are shown in Section 3. Extensive experimental results demonstrate the effectiveness of the proposed method compared to other SOTA methods on the benchmark datasets.

The contributions of this paper are summarized below:
\begin{itemize}
	\item We propose a novel non-uniform motion deblurring method with Atrous Spatial Pyramid Deformable Convolution (ASPDC) modules that realize region-specific convolution and different sizes of receptive fields, simultaneously. Our ablation study shows that both of these features significantly improve the performance using the same baseline.
	\item Different dilation rates in the ASDPC module extract information of different magnitudes of motion and separate the image into regions with the help of attention maps, which reduces the workload of each branch by focusing on regions with specific magnitudes of motion, instead of always considering the entire image.
	\item Our proposed deblurring network is end-to-end and contains only a single stage, which is more efficient and has a lower computational cost compared with other multi-stage methods.
	\item We propose a new end-to-end reblurring network that maps images of the deblurred domain back to the blurred domain. Then the deblurred output can be further refined using a new deblurring-reblurring consistency loss without estimating motion information. 
\end{itemize}

\begin{figure*}[t]
	\centering
	\includegraphics[width=0.7\textwidth]{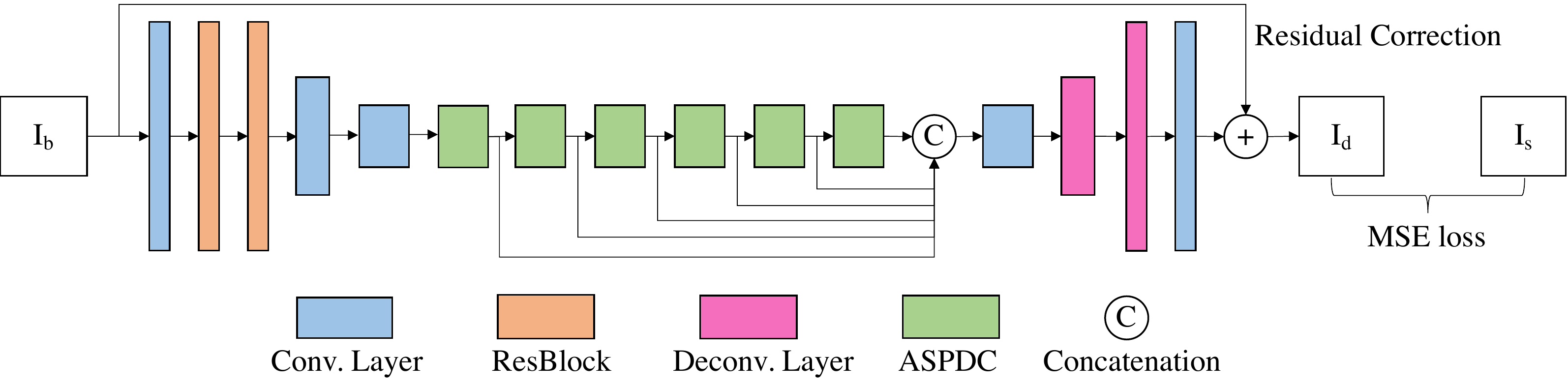}
	\caption{Overview of the deblurring network architecture.}
	\label{fig:deblurring_architecture}
\end{figure*}
\begin{figure}
	\centering
	\begin{subfigure}[h]{0.45\textwidth}
		\includegraphics[width=\textwidth]{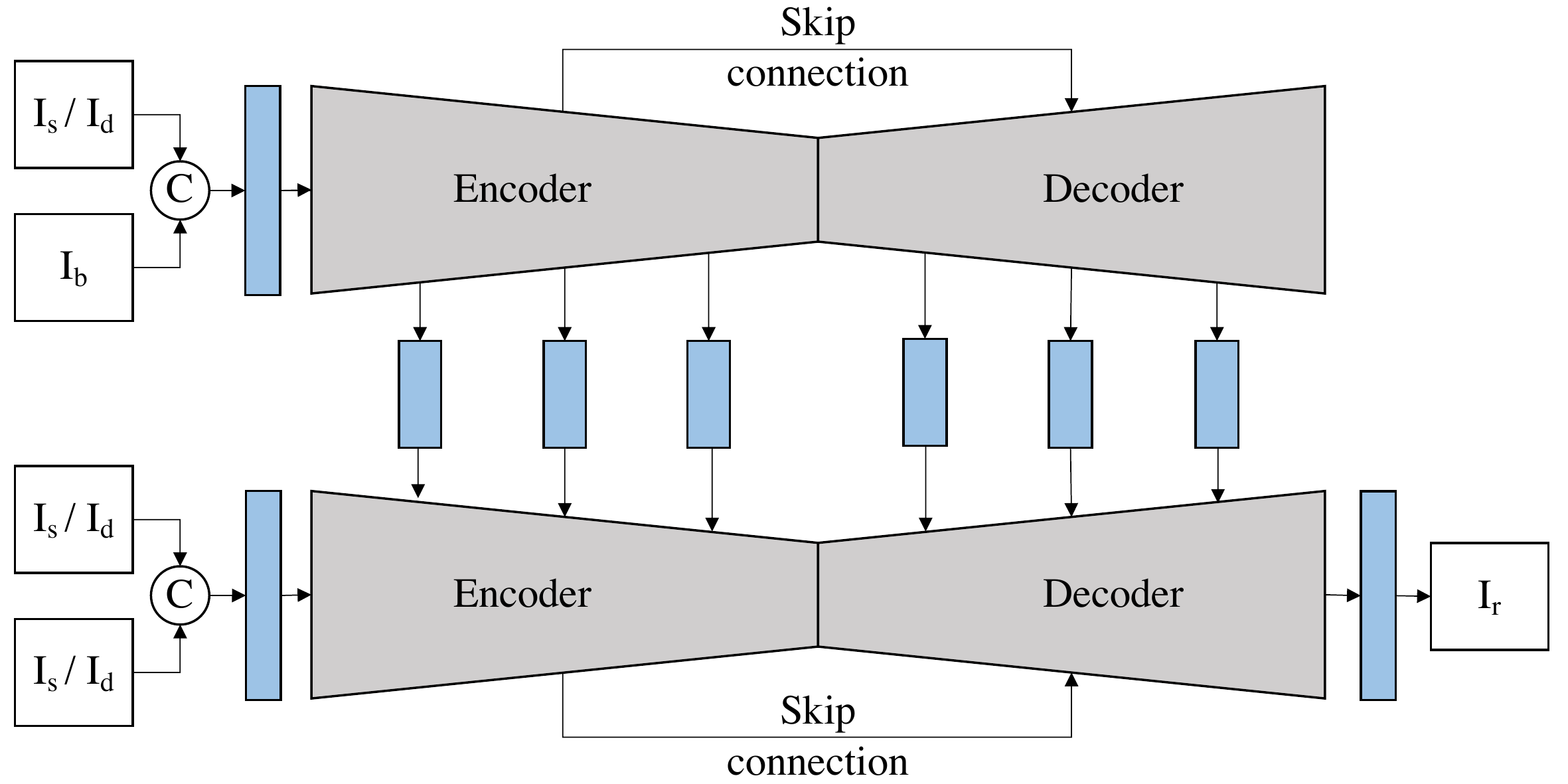}
		\subcaption{Reblurring network}
		\label{fig:reblurring}
	\end{subfigure}
	\begin{subfigure}[h]{0.45\textwidth}
		\includegraphics[width=\textwidth]{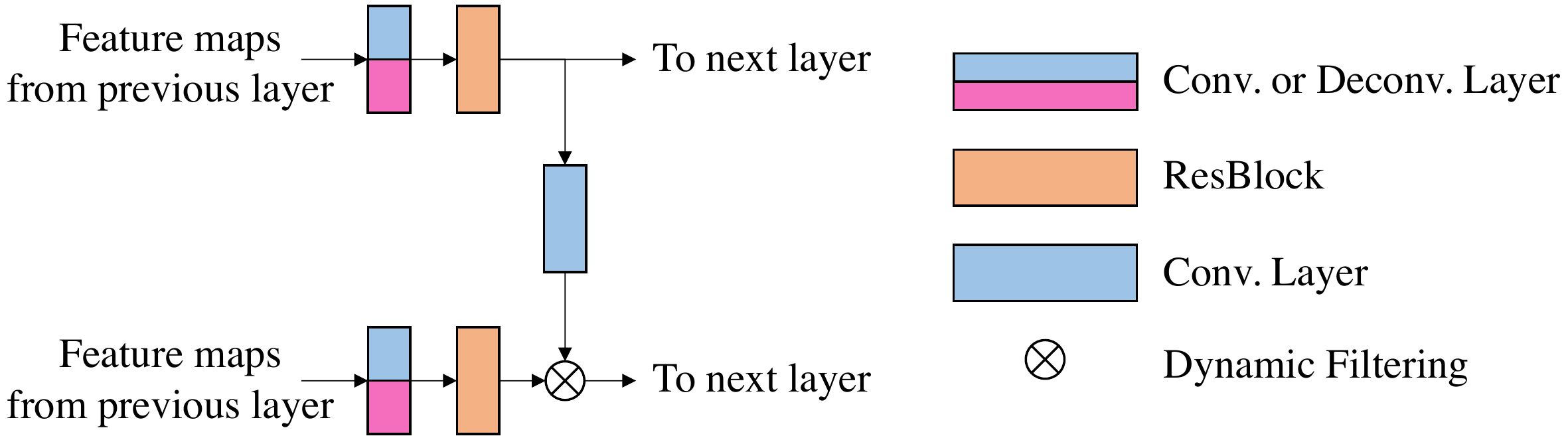}
		\subcaption{Feed blurred information from the upper to lower branch}
		\label{fig:dyna_filtering}
	\end{subfigure}
	\caption{Overview of the reblurring network architecture.}
	\label{fig:reblurring_architecture}
\end{figure}
\begin{figure}[t]
	\centering
	\begin{subfigure}[h]{0.4\textwidth}
		\includegraphics[width=\textwidth]{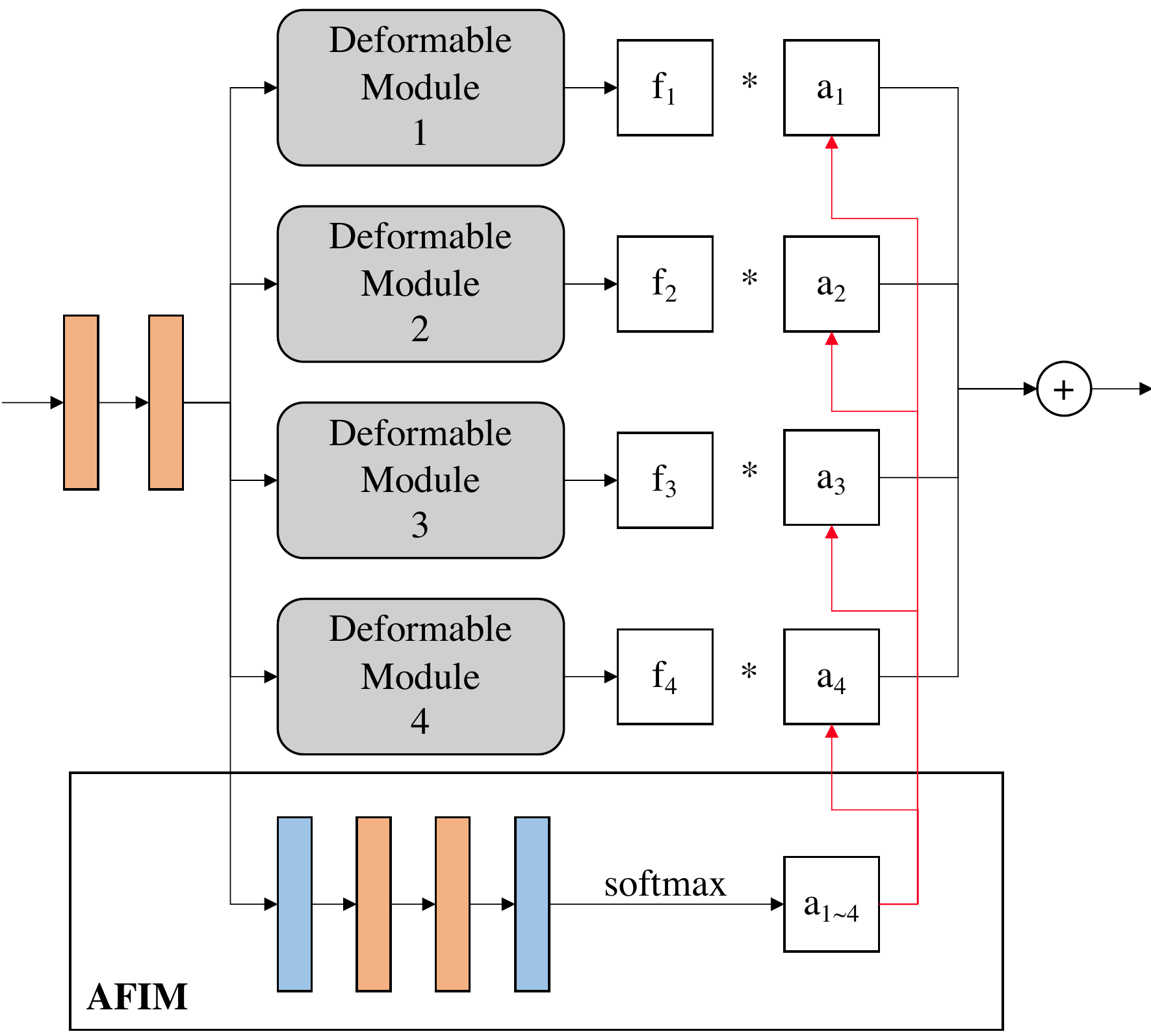}
	\end{subfigure}
	\begin{subfigure}[h]{0.38\textwidth}
		\includegraphics[width=\textwidth]{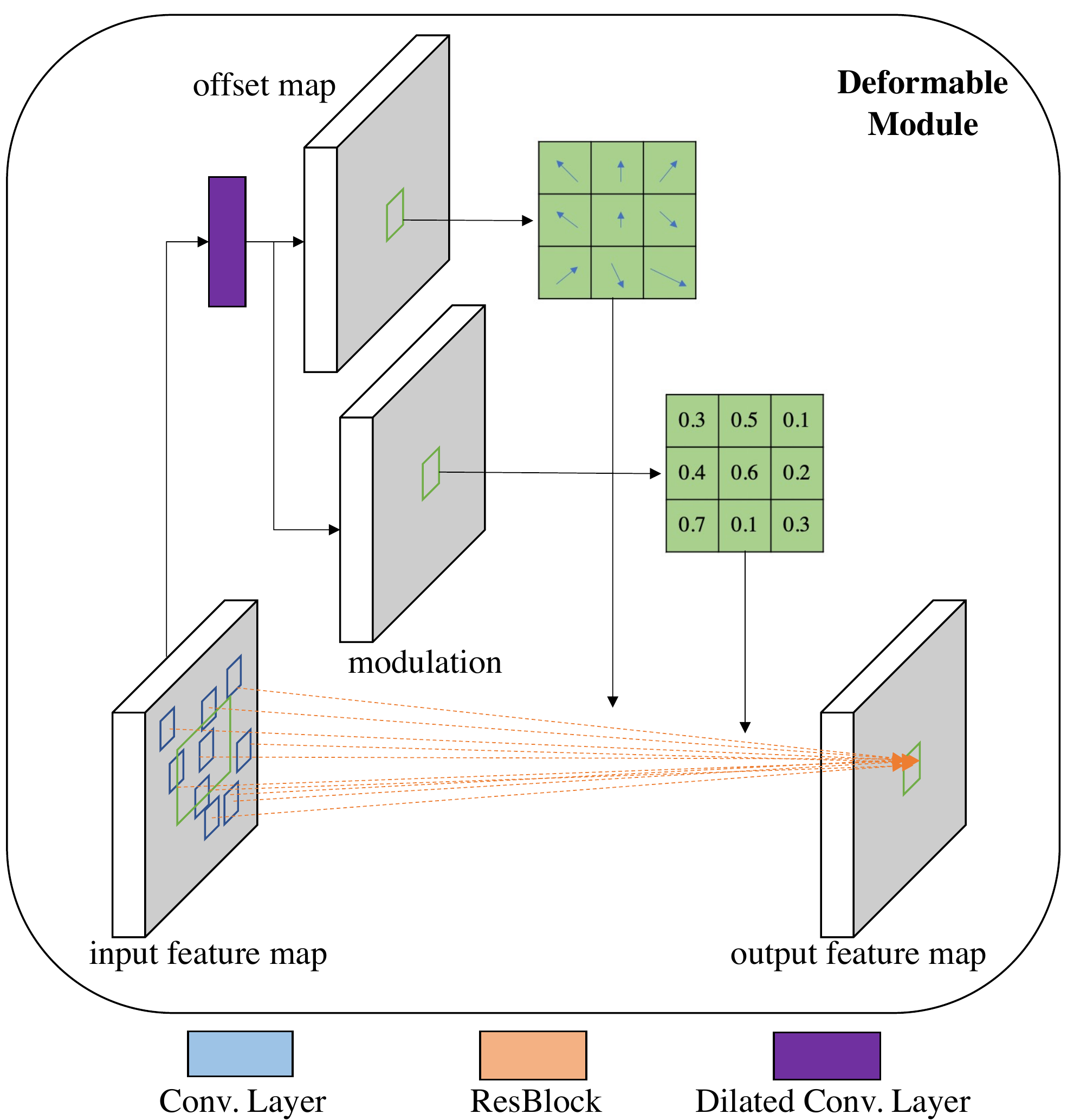}
	\end{subfigure}
	\caption{Schematic of the ASPDC module.}
	\label{fig:ASPDC}
\end{figure}

\section{Related Work}
\subsection{Blind Non-uniform Image Deblurring}
Since the blur kernel is spatially variant and unknown, blind non-uniform image deblurring is an ill-posed problem. Conventional methods usually constrain the condition of the blur and the image. Jia~\cite{jia2007single} assumes that the image contains only one moving object near the image center, and the transparency region of the blurred image helps to calculate the upper bound of the blur kernel. Gupta \etal~\cite{gupta2010single} calculate the integrated camera pose of selected patches with motion density function to estimate the local blur kernel and deblur the fronto-parallel scene. Hyun \etal~\cite{hyun2013dynamic} segment the image and handle the objects and background separately. Hyun \etal~\cite{hyun2014segmentation} utilize the edge-aware regularization to make the motion flow of neighboring pixels similar and to preserve edges. Anwar \etal~\cite{anwar2015class} explore a class-specific prior to deblur a specific kind of object. Bahat \etal~\cite{bahat2017non} find that when the blur kernel is an ideal low pass filter, applying the blur kernel to a blurred image will not change it. However, if the kernel is non-ideal, adding pink noise is helpful to estimate the blur kernel. The sharp image is recovered using the patch recurrence property~\cite{michaeli2014blind} within and across scales with the estimated local kernel. Bai \etal~\cite{bai2019single} prove that the extreme downsampled case of a blurred image is an approximation of the latent sharp image, which can be used as the prior image to reconstruct the latent sharp image from coarse to fine.

In order to improve the generalization of deblurring, deep learning based methods, such as a Convolutional Neural Network (CNN), are applied to solve this problem. Some methods focus on estimating motion information to use it for recovering the latent sharp image. Sun \etal~\cite{sun2015learning} combine the conventional methods and CNN. They assume that the blur kernel is locally invariant and the motion is linear, and use a CNN to estimate the motion field of overlapping patches. Then a Markov Random Field (MRF) model~\cite{li1994markov} is used to fuse the recovered patches. Gong \etal~\cite{gong2017motion} estimate the motion vector of each pixel and recover the whole sharp image directly instead of individual patches, which enables taking advantage of the context information of the image. Thekke \etal~\cite{tm2018semi} attempt to learn the weights of camera poses from a set of poses. Then the latent sharp image is reblurred with the estimated weights to calculate the reblurring error. Such a strategy can keep the cycle consistency, but it focuses on inplane motion only and the consistency is limited by a predefined camera pose set. In contrast, ours can recover more general blur without putting any specific limitations.

Some methods directly restore the sharp image in one stage. DeblurGAN~\cite{kupyn2018deblurgan} has an end-to-end architecture with multiple residual blocks~\cite{he2016deep} and instance normalization layers~\cite{ulyanov2016instance}. The authors use WGAN-GP~\cite{gulrajani2017improved} to stabilize the adversarial training. DeblurGAN-v2~\cite{kupyn2019deblurgan} extends the previous work~\cite{kupyn2018deblurgan} using a much deeper architecture with an encoder-decoder network. Since the non-uniform deblurring can be formulated using the Infinite Impulse Response (IIR) model~\cite{liu2016learning}, Zhang \etal~\cite{zhang2018dynamic} exploit the spatially variant RNN to take advantage of the long-range dependency in four directions. Aljadaany \etal~\cite{aljadaany2019douglas} implement the Douglas-Rachford splitting method~\cite{eckstein1992douglas} in a CNN which combines the advantages of both optimization and deep learning. Zhang \etal~\cite{zhang2020deblurring} attempt to train the model on realistic blurred images generated from a GAN. Yuan \etal~\cite{yuan2020efficient} utilize the Dense Inverse Search algorithm~\cite{kroeger2016fast} to estimate optical flow, then use it to guide the offset map in deformable convolution v2~\cite{zhu2019deformable}, which combines motion estimation with end-to-end learning. 

Some methods restore the sharp image in multiple stages. Nah \etal~\cite{nah2017deep} propose a multi-scale architecture to deblur the dynamic scene from coarse to fine. Tao \etal~\cite{tao2018scale} share the parameters of different scales. Zhang \etal~\cite{zhang2019deep} and Suin \etal~\cite{suin2020spatially} crop the blurred image into multiple patches of different sizes instead of downsampling to multiple scales. Thus, they preserve high frequency information. Park \etal~\cite{park2019multi} use a multi-temporal strategy to gradually sharpen the output.

\subsection{Spatially Variant Convolution}
Dai \etal~\cite{dai2017deformable} propose to learn an offset map to change the shape of the convolution filters, the result of which is more robust to geometric transformations. Huo \etal~\cite{huo2020blind} simplify the offset map of Dai \etal~\cite{dai2017deformable} by learning an offset for each pixel instead of each region but with more maps in a single layer. Zhu \etal~\cite{zhu2019deformable} extend the work of Dai \etal~\cite{dai2017deformable} by learning an extra modulation to weight the convolution filter, which can refine the effective receptive fields. Jia \etal~\cite{jia2016dynamic} use an extra branch to learn the filter dynamically, but the computational cost is high for large feature maps. Ding \etal~\cite{ding2019semantic} preserve the semantic correlation by multiple paired convolutions in local patches. The pixel-adaptive convolution of Su \etal~\cite{su2019pixel} can be regarded as a simplified version of the paired convolution~\cite{ding2019semantic} by reducing the kernel size to one. Dai \etal ~\cite{dai2020convolutional} exploit the position information of each pixel. The local patches extracted by the data kernel are concatenated with their positions in the image, and then are passed to a convolution layer.

\section{Proposed Method}
\subsection{Overview}

Overviews of the proposed deblurring and reblurring network architecture are illustrated in Figure~\ref{fig:deblurring_architecture} and Figure~\ref{fig:reblurring_architecture}, respectively. To make the training more stable, we initially train the two networks separately. The deblurring network attempts to recover the sharp image $I_s$ from the blurred input $I_b$, then the deblurred output $I_d$ is reblurred by the reblurring network. The reblurring network takes both $I_s$ and $I_b$ as input and outputs the reblurred image $I_r$. When the training of two networks are converged, we replace the input $I_s$ of the reblurring network with the deblurred output $I_d$, and fine-tune the $I_d$ with the deblurring-reblurring consistency loss. We do not use any kind of normalization (e.g. batch normalization~\cite{ioffe2015batch} or instance normalization~\cite{ulyanov2016instance}).

\begin{table*}[t]
\centering
\begin{tabular}{|c|c|c|c|c|c|c|c|c|}
\hline
Method & Xu~\cite{xu2013unnatural}    & Sun~\cite{sun2015learning} & Nah~\cite{nah2017deep}   & Kupyn~\cite{kupyn2018deblurgan} & Tao~\cite{tao2018scale}    & Zhang~\cite{zhang2018dynamic} & Kupyn~\cite{kupyn2019deblurgan}        & Aljadaan~\cite{aljadaany2019douglas} \\ \hline
PSNR   & 20.30 & 25.31 & 28.49 & 28.70 & 30.26   & 29.19 & 29.55        & 30.35    \\ \hline
SSIM   & 0.741 & 0.851 & 0.917 & 0.927 & 0.934   & 0.931 & 0.934        & \textcolor{red}{0.961}    \\ \hline\hline
Method & Zhang~\cite{zhang2019deep} & Suin~\cite{suin2020spatially}  & Park~\cite{park2019multi}     & Yuan~\cite{yuan2020efficient}  & Purohit~\cite{purohit2020region} & Zhang~\cite{zhang2020deblurring} & Ours & Ours+     \\ \hline
PSNR   & 31.20 & \textcolor{blue}{32.02} & 31.15  & 29.81 & 31.76   & 31.10 &  31.97      &  \textcolor{red}{32.09}        \\ \hline
SSIM   & 0.945 & 0.953 & 0.945 & 0.937 & 0.953   & 0.942 &  0.957       &  \textcolor{blue}{0.959}        \\ \hline
\end{tabular}
\caption{Quantitative comparison on the GoPro dataset~\cite{nah2017deep}. Ours/Ours+ represents our deblurring network without/with fine-tuning on the reblurring network. The best results are in \textcolor{red}{red} and the second best in \textcolor{blue}{blue}.}
\label{tab:comparison_1}
\end{table*}
\begin{table*}[t]
\centering
\begin{tabular}{|c|c|c|c|c|c|c|c|c|c|}
\hline
Method & Kupyn~\cite{kupyn2018deblurgan} & Tao~\cite{tao2018scale}   & Zhang~\cite{zhang2019deep} & Park~\cite{park2019multi} & Suin~\cite{suin2020spatially}  & Shen~\cite{shen2019human} & Kupyn~\cite{kupyn2019deblurgan} & Ours  & Ours+ \\ \hline
PSNR   & 24.51 & 28.36 & 29.09 & 29.16 & \textcolor{blue}{29.98} & 28.89 & 26.61 & \textcolor{blue}{29.98} & \textcolor{red}{30.04} \\ \hline
SSIM   & 0.871 & 0.915 & 0.924 & 0.933 & 0.930 & 0.930 & 0.875 & \textcolor{blue}{0.944} & \textcolor{red}{0.945} \\ \hline
\end{tabular}
\caption{Quantitative comparison on the HIDE dataset~\cite{shen2019human}. Ours/Ours+ represents our deblurring network without/with fine-tuning on the reblurring network. The best results are in \textcolor{red}{red} and the second best in \textcolor{blue}{blue}.}
\label{tab:comparison_2}
\end{table*}

Our deblurring-reblurring consistency is inspired by Guo \etal~\cite{guo2020closed}. Deblurring and reblurring can be regarded as a pair of dual tasks. The former is the primary task and the latter is the corresponding dual task, which is similar to upsampling and downsampling in super-resolution~\cite{yang2019deep}. Guo \etal~\cite{guo2020closed} proves that the generalization bound of the dual regression (in our case, consistency) is lower than that of the primary regression (only deblurring). Therefore, it leads to more accurate deblurring results. However, simply mapping the deblurred output to the original blurred input is highly ill-posed. Lu \etal~\cite{lu2019unsupervised} show that the sharp image only contains the sharp content without any blur information, so it needs the blur information from the corresponding blurred image as the extra input for reblurring. Thus, we also utilize blur information.

\subsection{Deblurring Network}
We use two residual blocks (ResBlocks)~\cite{he2016deep} and strided convolution layers to extract high dimensional features at the beginning, and two deconvolution layers to recover the spatial dimension in the end. The last convolution layer reduces the channel size of the feature map to 3 (RGB). We find that learning the residual correction instead of directly learning the latent sharp image can make the training more stable and faster. To explain our intuition, note that the blurred image contains all of the signals from the sensor during the exposure time, as shown in Eqn~\ref{eqn:integration_model}. One of the sharp images $I_{S(t)}$, say at time $t^*$, is the corresponding target while many features extracted from $I_b$ could be from times other than $t^*$. Hence, learning features of the other times will be easier than directly learning a specific one. 

In the middle of the deblurring network, we stack six ASPDC modules (as shown in Figure~\ref{fig:ASPDC}) in which we extend the work of the deformable convolution network v2 (DCNv2)~\cite{zhu2019deformable}. The original DCNv2 applies different convolution kernels to different regions by learning an offset map $\Delta p$ and a modulation $\Delta m$. But a fixed size is used for the receptive field of each region used to generate $\Delta p$ and $\Delta m$. 
Since $\Delta p$ represents the shift of each pixel, it can be regarded as the local optical flow~\cite{yuan2020efficient} corresponding to the motion of the object and the camera. For a non-uniform blurred image, some of the regions might have only small variations while other regions might have large movements and overlaps. In this case, the original DCNv2 uses a single convolution layer to generate $\Delta p$ and $\Delta m$ and treats these regions similarly, which is not an optimal choice for this problem. 

To make the deformable convolution more flexible, in our ASPDC module, we build four branches with different dilation rates~\cite{yu2015multi} to generate four offset maps $\Delta p$ and modulations $\Delta m$ with different receptive fields, and four deformable convolution outputs. As shown in Figure~\ref{fig:ASPDC}, the dilation rates of dilated convolution layers in deformable modules 2$\sim$4 are 1, 2, and 4, respectively. The deformable module 1 also uses the dilation rate 1 but it ignores the offsets (by setting $\Delta p$ as zero). Such a special module is used to recover static regions.

The outputs of the four branches in an ASPDC module are fused by an attention feature integration module (AFIM)~\cite{deng2019deep}. We can write it as:
\begin{gather}
f_o= \sum_{i = 1}^{4} a_i * f_i,\\
\sum_{i = 1}^{4} a_{ij} = 1, 1 \leq j \leq h \times w,
\label{eqn:weighted_sum}
\end{gather}  
in which $f_i$ is the output of the $i^{th}$ branch, $a_i$ is a single-channel attention map generated from the AFIM, $j$ is the index of the pixel, $h$ and $w$ are, respectively, the height and width of the attention map, $*$ represents the element-wise multiplication and $f_o$ is the output of the ASPDC module. To make sure that the channel-wise sum of attention maps is 1, we utilize the softmax activation along the channel. In this case, each region that integrates information from different receptive fields significantly boosts the performance. The output feature maps of six ASPDC modules are further concatenated to stabilize training.

We use the Mean Squared Error (MSE) loss as our final deblurring loss:
\begin{gather}
L_{deblurring} = || I_s - I_d ||^2_F,
\end{gather}  
where $I_s$ and $I_d$ are the sharp target and the deblurred output, respectively.

\subsection{Reblurring Network}
In order to narrow down the solution space of deblurring and to refine the deblurred output $I_d$, we build an end-to-end reblurring network to reblur the deblurred output and calculate the deblurring-reblurring consistency loss. Simply mapping the sharp image back to the blurred image is not impossible but difficult. Because the non-uniform blurred image domain is much larger than the sharp image domain. Hence, we need blur information from the blurred image to assist the mapping. However, directly inputting sharp and blurred image together and outputting the reblurred image is difficult to train, since the training procedure is unstable and easy to collapse to an identity mapping. The network may choose to output the blurred input directly and ignore the sharp input, which is definitely undesirable.

To handle the above problem, we utilize an architecture which is able to take full use of blur information and avoid training collapse. As shown in Figure~\ref{fig:reblurring}, the network contains two encoder-decoder branches, and the weights of the two branches are shared for reducing the number of parameters. The architecture of the encoder-decoder is simple, which consists of multiple conv/deconv-resblock pairs (convolution layers for the encoder and deconvolution layers for the decoder) as shown in Figure~\ref{fig:dyna_filtering}. The concatenation of blurred and sharp image is used as the input of the upper branch, and two duplicate sharp images are input into the lower branch for matching the channel dimension. 

The upper branch learns to compare the blurred and the sharp image and passes feature maps with blur information to the lower branch after each conv/deconv-resblock pair. In the lower branch, a convolution layer reduces channels of feature maps from the upper branch into $K\times K$ to generate a dynamic local filter~\cite{jia2016dynamic} for each pixel. For reducing computational cost of dynamic filtering, we set $K$ as 3 and apply the same filter on all channels of feature maps from the lower branch. The dynamic local filters are regarded as the spatial-variant blur kernels which gradually reblur the feature maps of the lower branch from beginning to end. The detailed architecture of the reblurring network is shown in supplementary material.

Similar to the deblurring network, we use the Mean Squared Error (MSE) loss here:
\begin{gather}
L_{reblurring} = || I_r - I_b ||^2_F.
\label{eqn:reblurring_loss}
\end{gather}

\begin{figure*}[t]
	\centering
	\begin{subfigure}[h]{0.34\textwidth}
		\includegraphics[width=\textwidth]{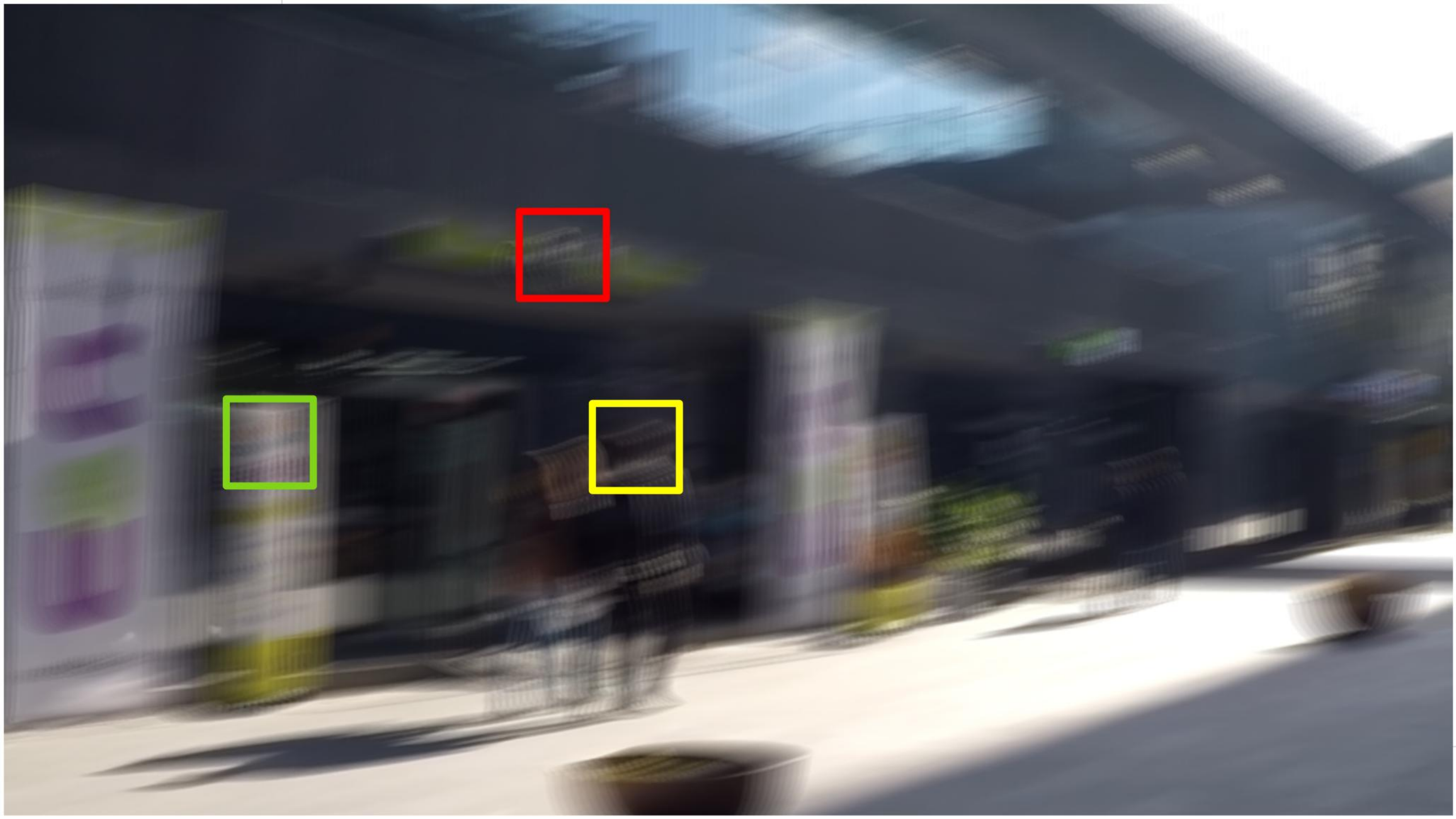}
		\caption{Blurred input}
	\end{subfigure}
	\begin{subfigure}[h]{0.1005\textwidth}
		\includegraphics[width=\textwidth]{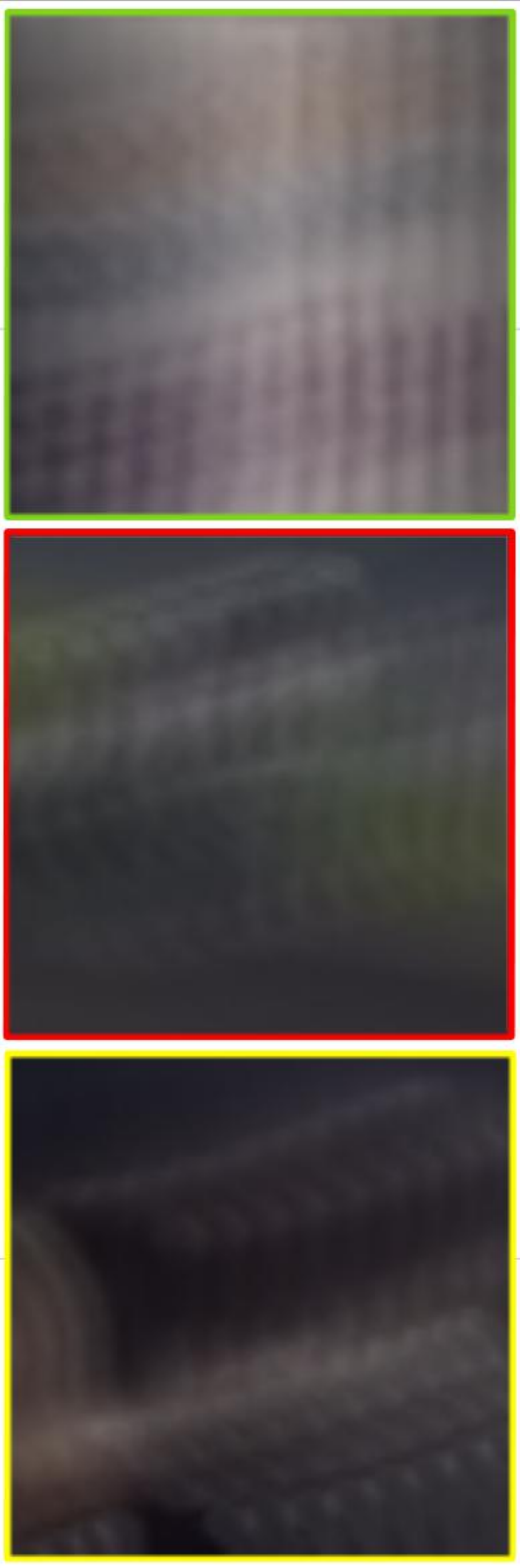}
		\caption{Blurred}
	\end{subfigure}
	\begin{subfigure}[h]{0.1005\textwidth}
		\includegraphics[width=\textwidth]{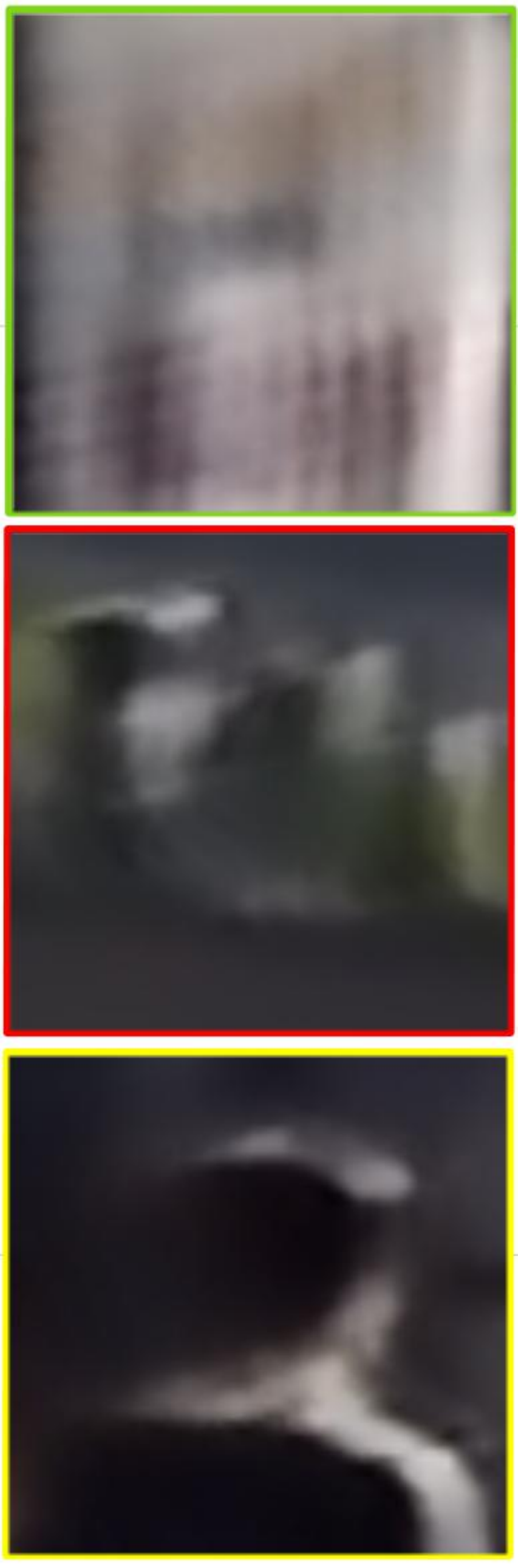}
		\caption{Park\cite{park2019multi}}
	\end{subfigure}
	\begin{subfigure}[h]{0.102\textwidth}
		\includegraphics[width=\textwidth]{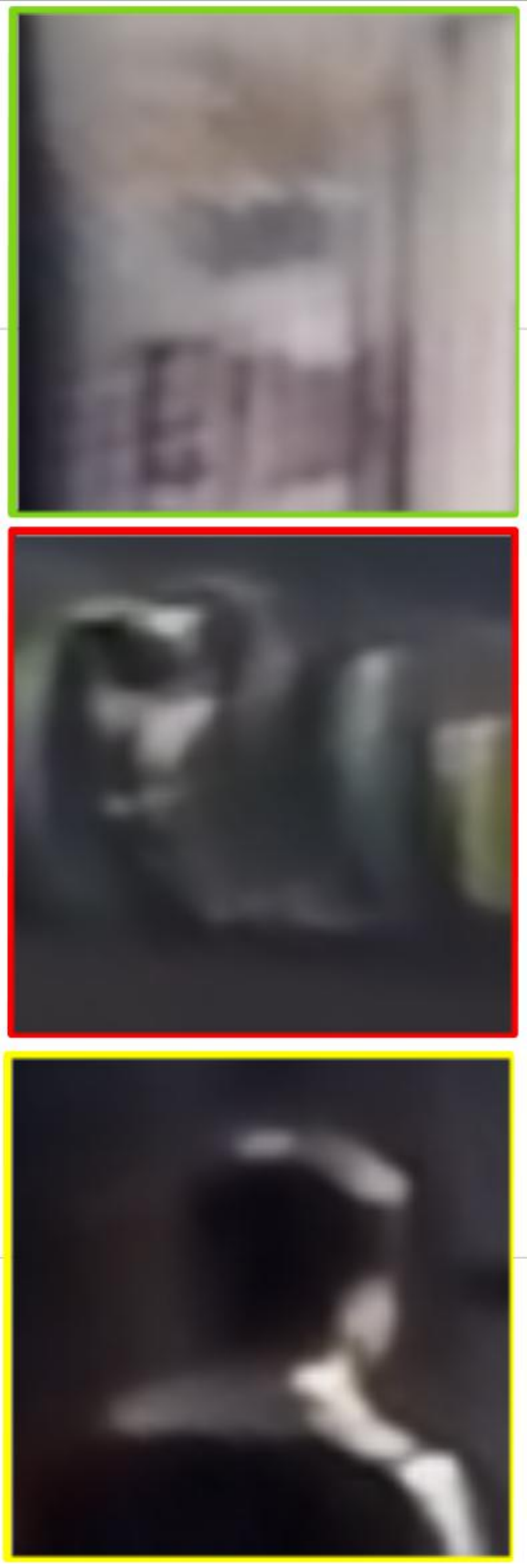}
		\caption{Zhang\cite{zhang2019deep}}
	\end{subfigure}
	\begin{subfigure}[h]{0.102\textwidth}
		\includegraphics[width=\textwidth]{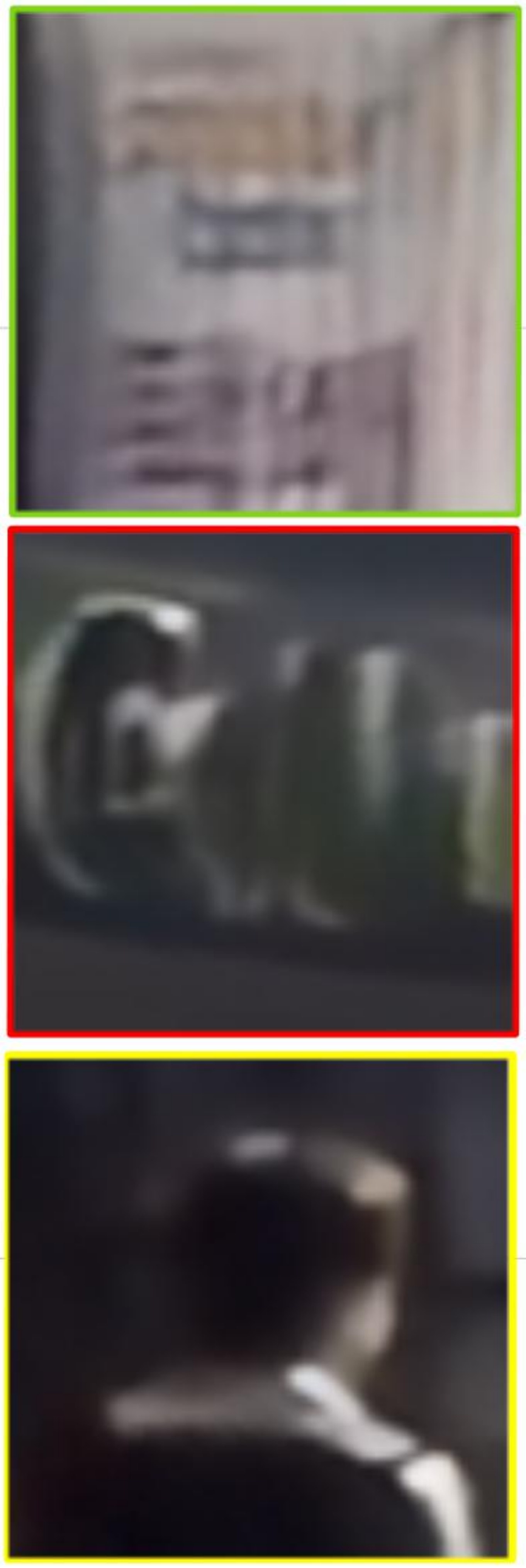}
		\caption{Prohit\cite{purohit2020region}}
	\end{subfigure}
	\begin{subfigure}[h]{0.101\textwidth}
		\includegraphics[width=\textwidth]{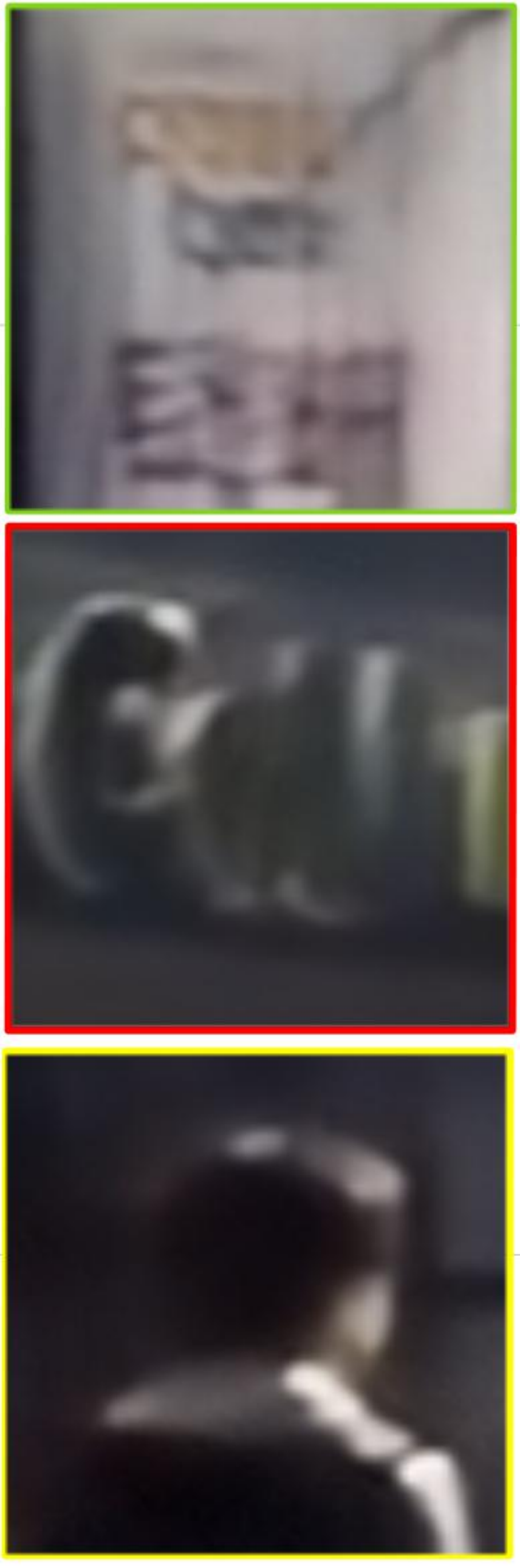}
		\caption{Ours+}
	\end{subfigure}
	\begin{subfigure}[h]{0.102\textwidth}
	\includegraphics[width=\textwidth]{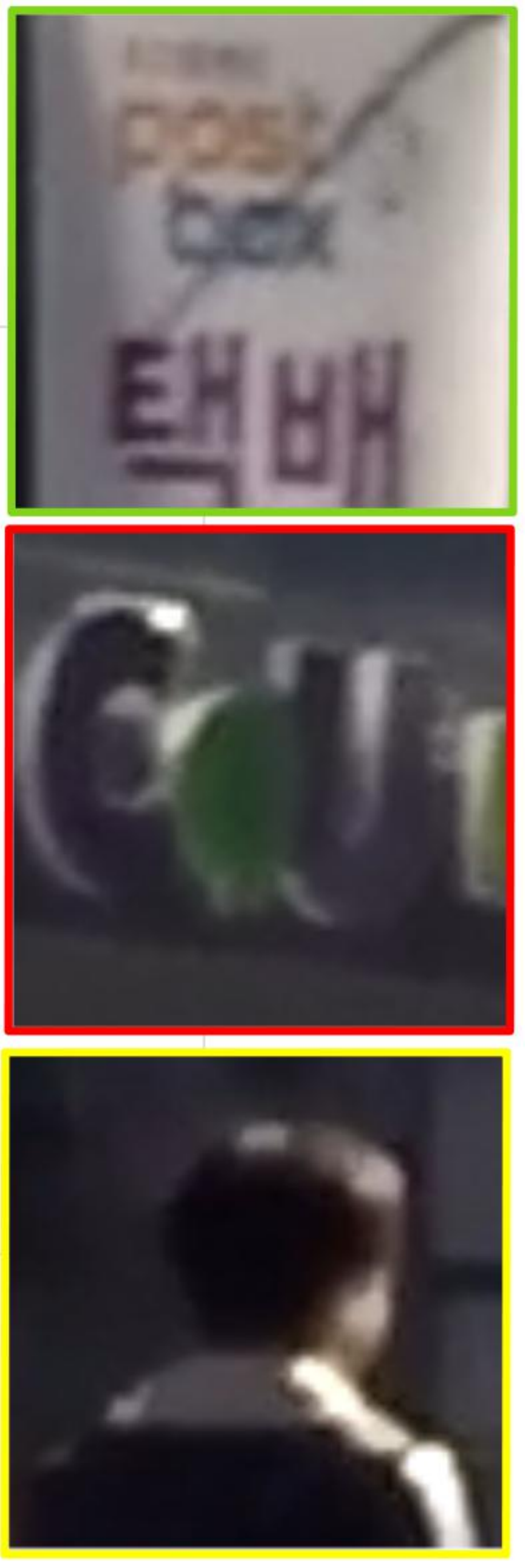}
	\caption{Sharp GT}
	\end{subfigure}
	\caption{Qualitative comparison on the GoPro dataset: a) Blurred input image. b-g) Magnified crops of the blurred input and deblurred outputs of compared methods, and the sharp ground truth.}
	\label{fig:qual_gorpo_main}
\end{figure*}
\begin{figure*}[t]
	\centering
	\begin{subfigure}[h]{0.19\textwidth}
		\includegraphics[width=\textwidth]{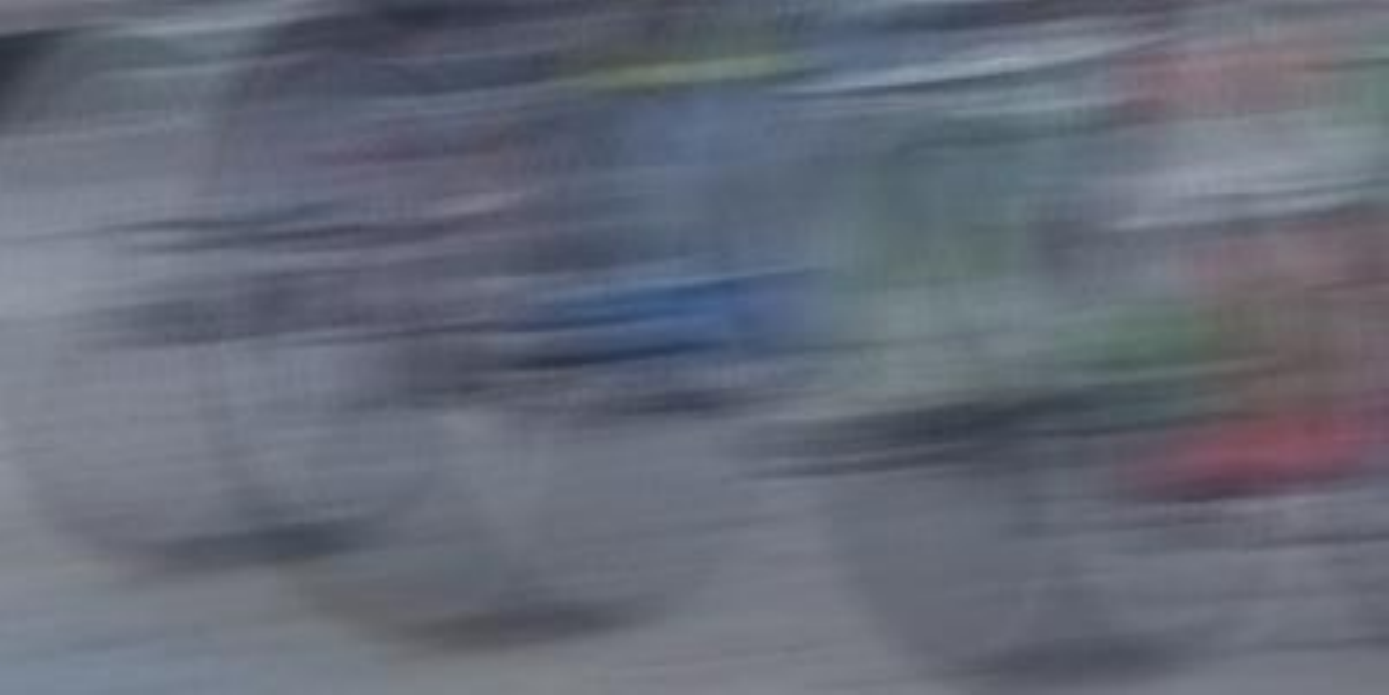}
		\caption{Blurred input}
	\end{subfigure}
	\begin{subfigure}[h]{0.19\textwidth}
		\includegraphics[width=\textwidth]{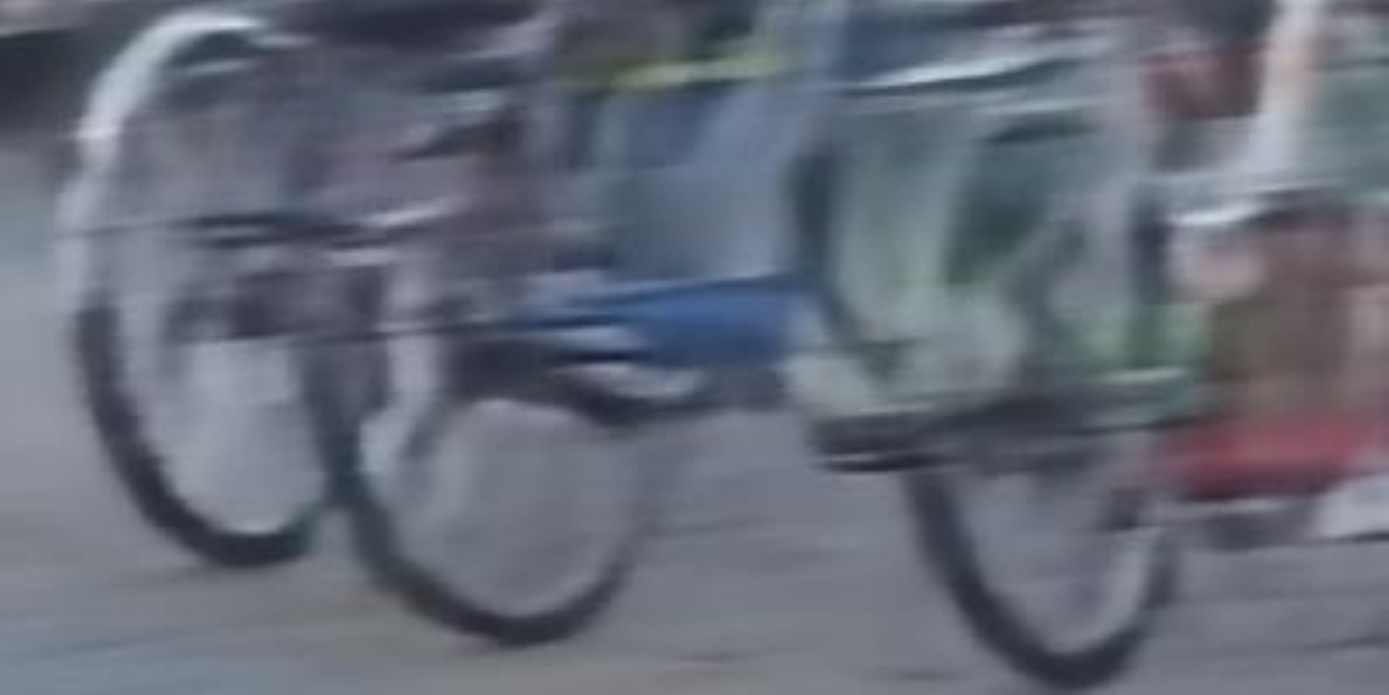}
		\caption{Park~\cite{park2019multi}}
	\end{subfigure}
	\begin{subfigure}[h]{0.19\textwidth}
		\includegraphics[width=\textwidth]{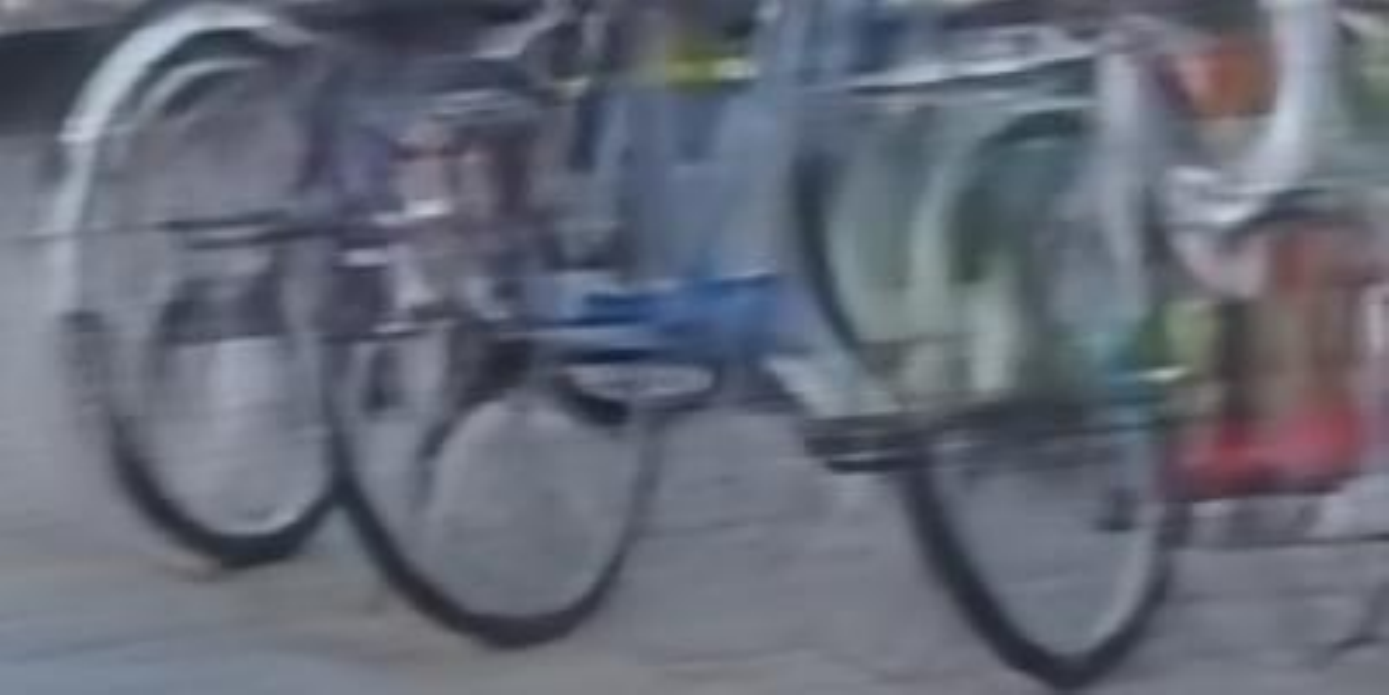}
		\caption{Zhang~\cite{zhang2019deep}}
	\end{subfigure}
	\begin{subfigure}[h]{0.19\textwidth}
		\includegraphics[width=\textwidth]{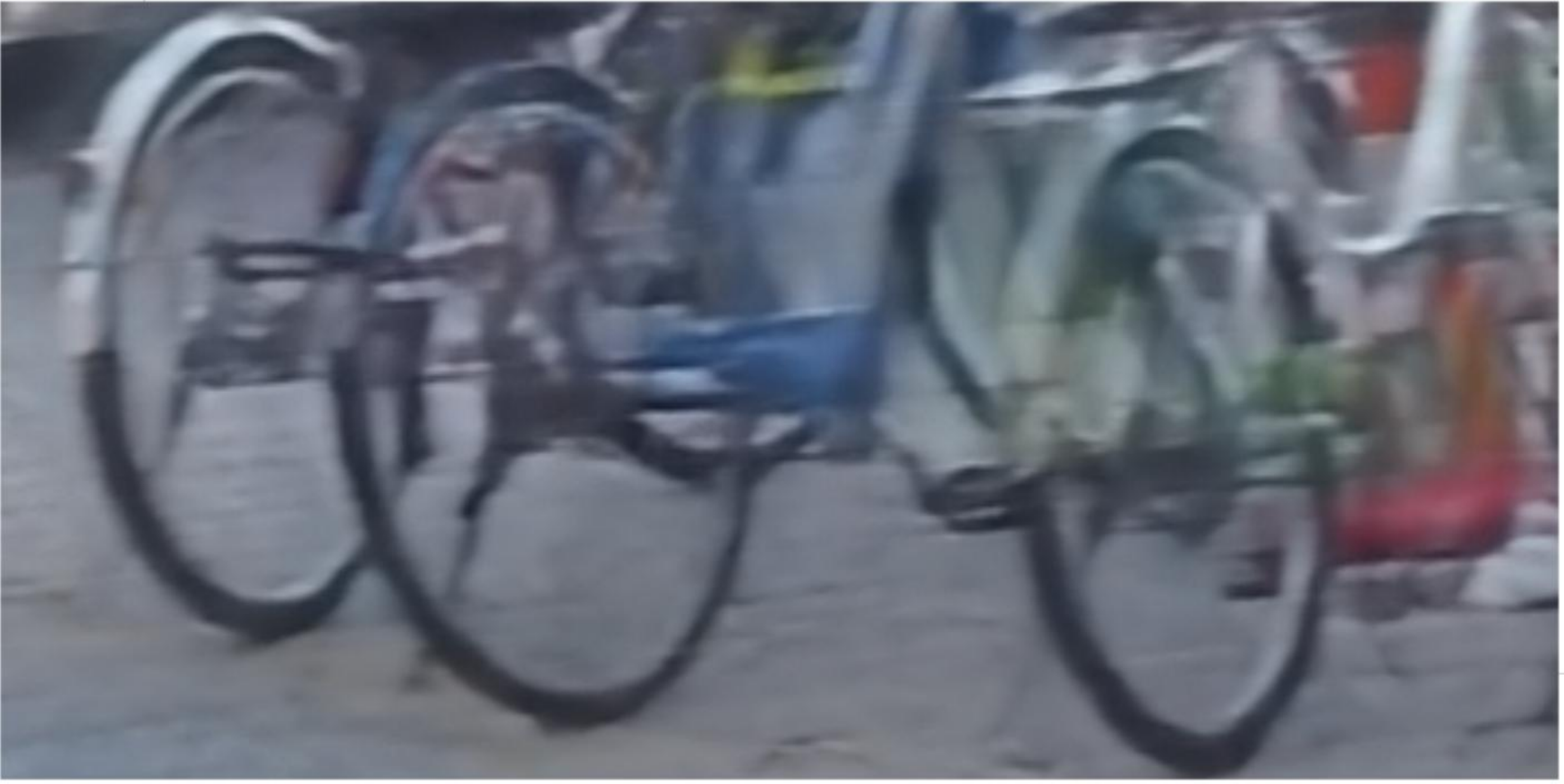}
		\caption{Ours+}
	\end{subfigure}
	\begin{subfigure}[h]{0.19\textwidth}
	\includegraphics[width=\textwidth]{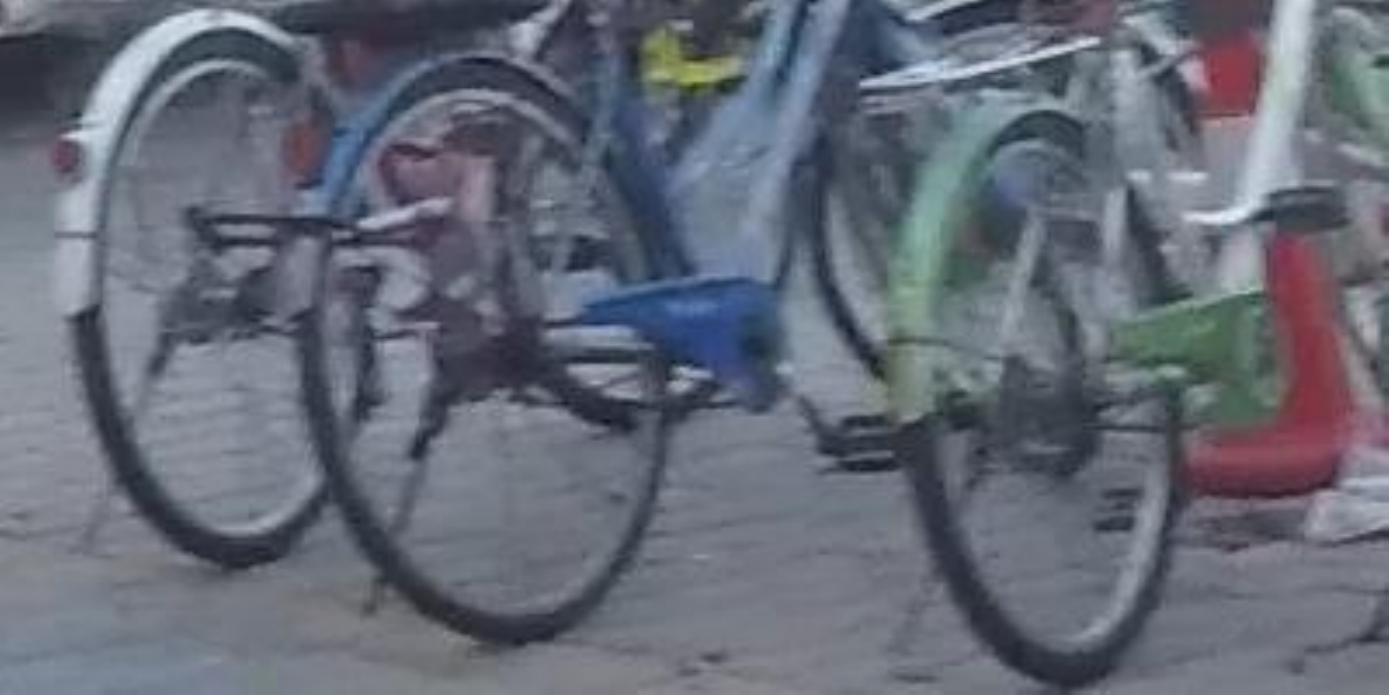}
	\caption{Sharp GT}
	\end{subfigure}
	\caption{Qualitative comparison on the HIDE dataset: a) Blurred input image. b-g) Magnified crops of the blurred input, deblurred outputs of compared methods, and the sharp ground truth.}
	\label{fig:qual_hide_main}
\end{figure*}

\begin{figure*}[h]
	\centering
	\begin{subfigure}[h]{0.4\textwidth}
		\includegraphics[width=\textwidth]{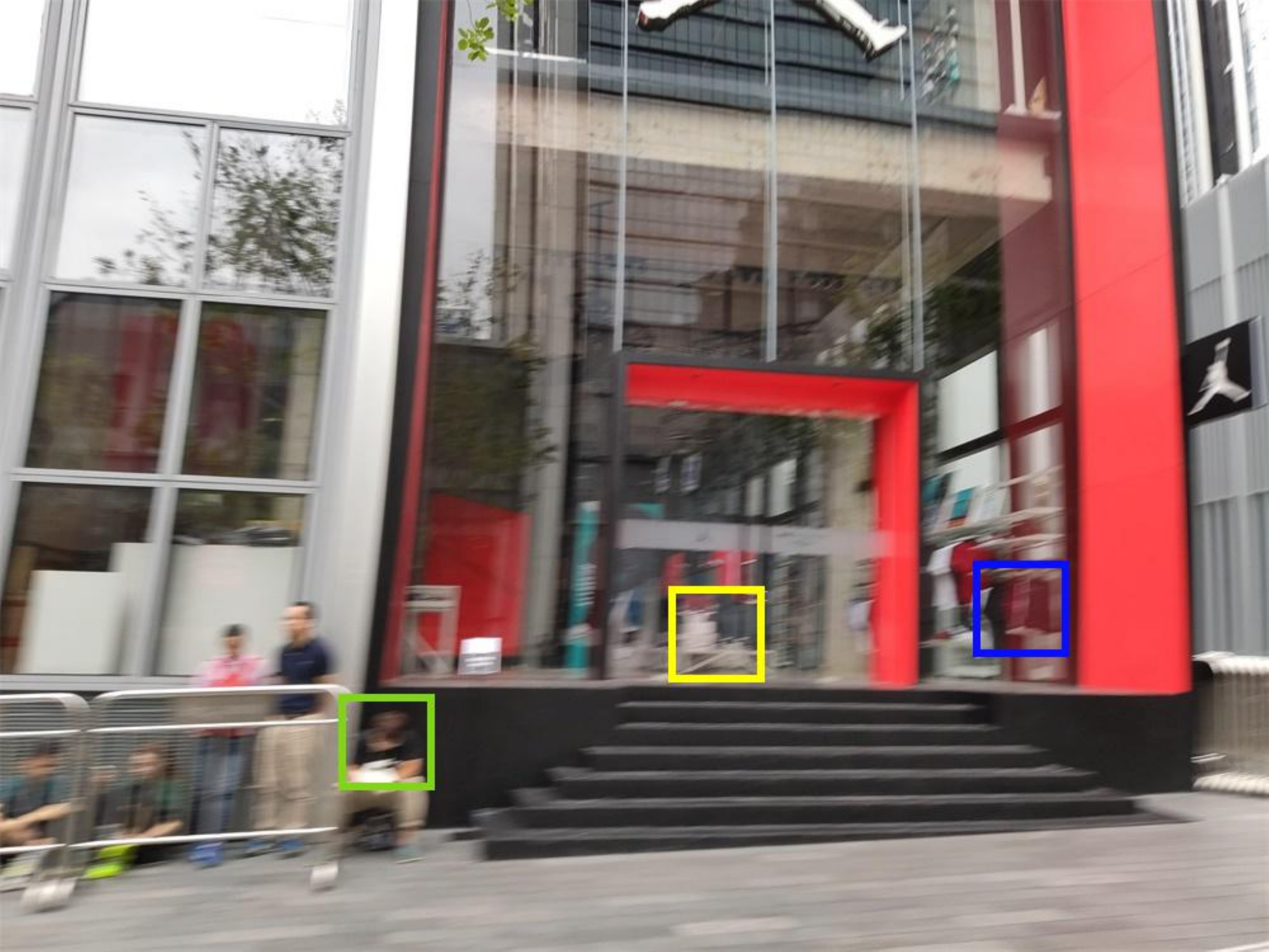}
		\caption{Blurred input}
	\end{subfigure}
	\begin{subfigure}[h]{0.1025\textwidth}
		\includegraphics[width=\textwidth]{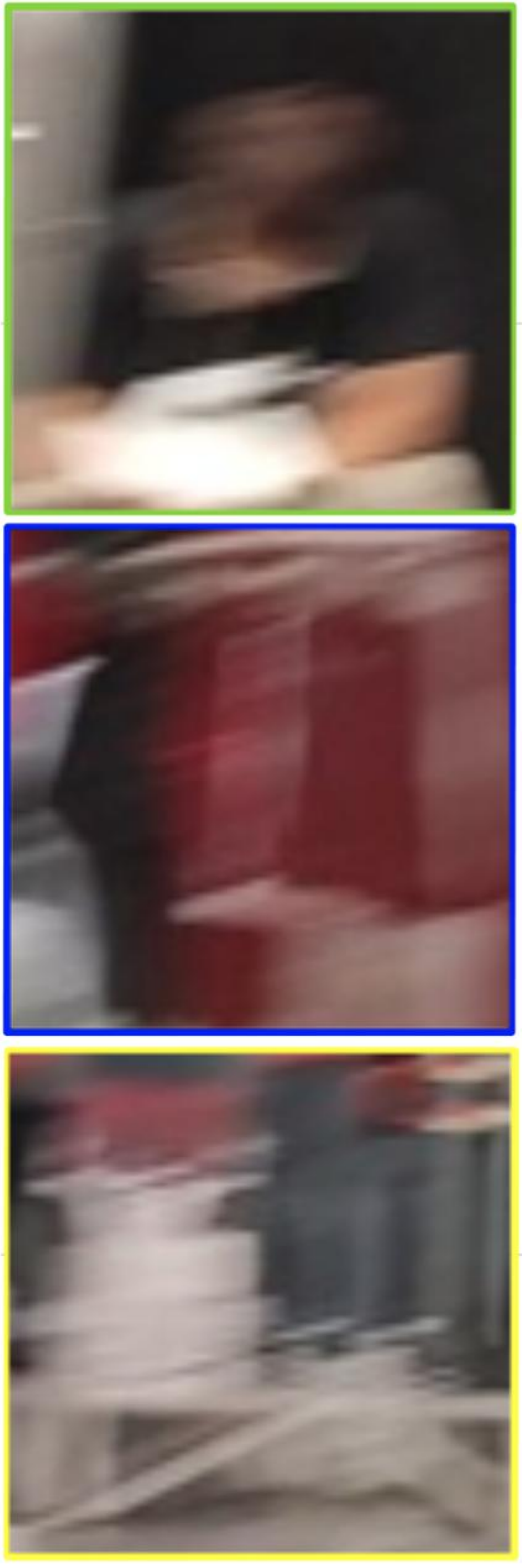}
		\caption{Blurred}
	\end{subfigure}
	\begin{subfigure}[h]{0.103\textwidth}
		\includegraphics[width=\textwidth]{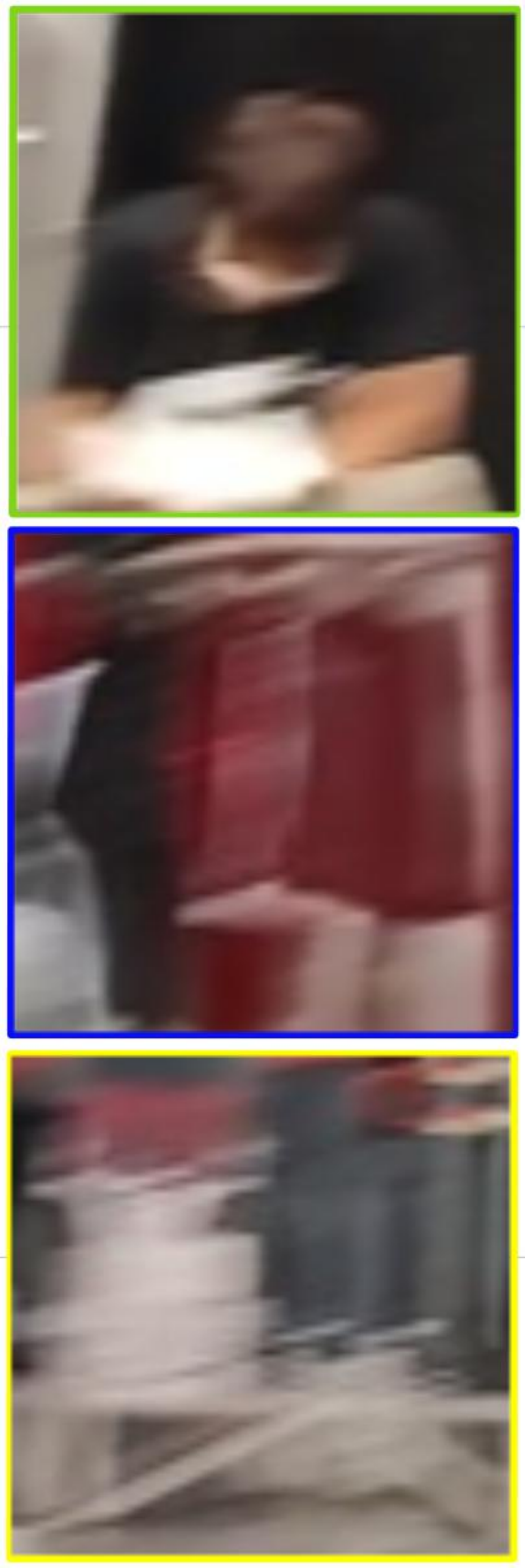}
		\caption{Park\cite{park2019multi}}
	\end{subfigure}
	\begin{subfigure}[h]{0.102\textwidth}
		\includegraphics[width=\textwidth]{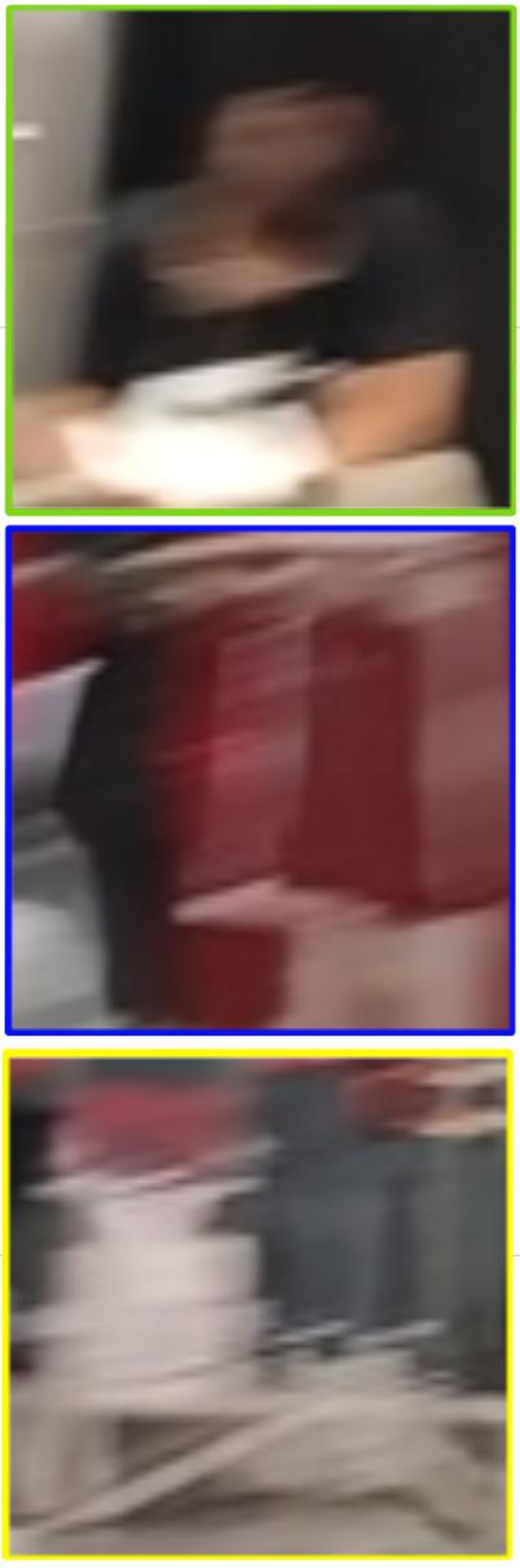}
		\caption{Zhang\cite{zhang2019deep}}
	\end{subfigure}
	\begin{subfigure}[h]{0.103\textwidth}
		\includegraphics[width=\textwidth]{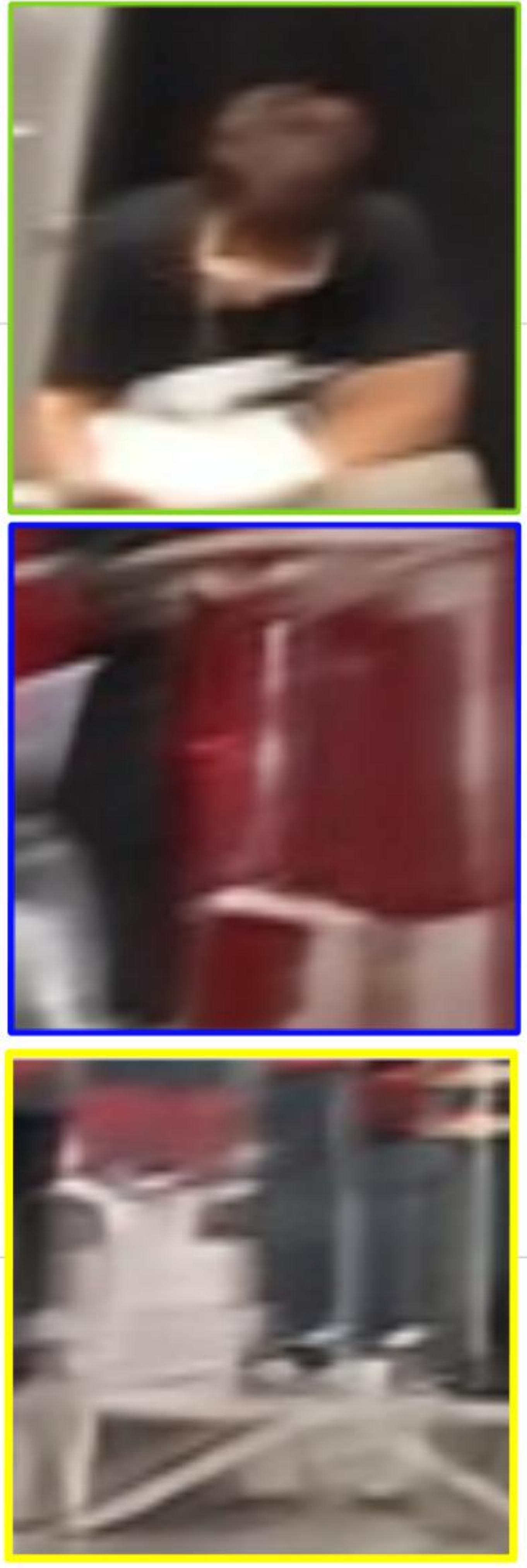}
		\caption{Ours+}
	\end{subfigure}
	\caption{Qualitative comparison on the RWBI dataset: a) The blurred input image. b-e) Magnified crops of the blurred input and the deblurred outputs of compared methods. Note that no ground truth is available for RWBI.}
	\label{fig:qual_rwbi_main}
\end{figure*}

\subsection{Fine-tuning}
After the training of the deblurring and reblurring network converges, we replace $I_s$ in the reblurring network with the deblurred output $I_d$ to refine $I_d$ by the deblurring-reblurring consistency loss. The loss function is defined as:
\begin{gather}
L_{consistency} =L_{deblurring} + \lambda L_{reblurring},
\label{eqn:overall_loss}
\end{gather}
where $\lambda$ is the weight of the reblurring loss, and we empirically set $\lambda = 0.1$.

\section{Experiments}
\subsection{Datasets}
We follow the literature~\cite{kupyn2018deblurgan, kupyn2019deblurgan, nah2017deep, suin2020spatially, zhang2019deep} to train our model on 2103 training images from the GoPro dataset~\cite{nah2017deep}. We then use 1111 testing images from the GoPro dataset and 2025 testing images from the HIDE dataset~\cite{shen2019human} as our testing set. We also do qualitative comparison on the Real World Blurred Image (RWBI) dataset~\cite{zhang2020deblurring}.

\subsection{Implementation Details}
The method is implemented in PyTorch~\cite{paszke2019pytorch} and evaluated on a single NVIDIA RTX 2080 Ti GPU. During training, we use Adam optimizer~\cite{kingma2014adam} with $\beta_1 = 0.9$, $\beta_2 = 0.999$ and $\epsilon = 10^{-18}$. All parameters are initialized using Xavier normalization~\cite{glorot2010understanding}. We randomly crop the training images into 256$\times$256 patch pairs and set the batch size as 6. The learning rate is initialized as $10^{-4}$ and halved every 1000 epochs. The training procedure is terminated when the learning rate reaches $10^{-6}$. During fine-tuning, we set the learning rate as $10^{-5}$ and it is halved every 200 epochs. We stop the fine-tuning when the learning rate reaches $10^{-6}$. The size of all convolution filters is 3$\times$3. We set the initial number of channels for all convolution layers and residual blocks to 32 in the deblurring network and 16 in the reblurring network, and we double (halve) them every time we downscale (upscale) the spatial dimension.

\subsection{Quantitative Comparison}
We first compare our method ("ours+") with some others on the 1111 testing images from the GoPro dataset, including a conventional method (Xu \etal~\cite{xu2013unnatural}), and some deep learning based methods (Sun \etal~\cite{sun2015learning}, Nah \etal~\cite{nah2017deep}, Kupyn \etal~\cite{kupyn2018deblurgan}, Tao \etal~\cite{tao2018scale}, Zhang \etal~\cite{zhang2018dynamic}, Kupyn \etal~\cite{kupyn2019deblurgan}, Aljadaany \etal~\cite{aljadaany2019douglas}, Zhang \etal~\cite{zhang2019deep}, Suin \etal~\cite{suin2020spatially}, Park \etal~\cite{park2019multi}, Yuan \etal ~\cite{yuan2020efficient}, Purohit \etal~\cite{purohit2020region} and Zhang \etal~\cite{zhang2020deblurring}). We also evaluate our deblurring network without fine-tuning on the reblurring network ("ours"). We use PSNR~\cite{NADIPALLY201921} and SSIM~\cite{wang2004image} as evaluation metrics. All the methods are trained on the GoPro dataset following the same strategy.

\begin{table}[t]
\centering
\begin{tabular}{|c|c|c|c|}
\hline
Method         & Kupyn~\cite{kupyn2019deblurgan} & Zhang~\cite{zhang2019deep} & Nah~\cite{nah2017deep}  \\ \hline
Time (sec)     & 1.68  & 0.40  & 0.93 \\ \hline
GPU (GB) & 2.41  & \textcolor{red}{2.10}  & 9.70 \\ \hline\hline
Method         & Tao~\cite{tao2018scale}   & Park~\cite{park2019multi}  & Ours \\ \hline
Time (sec)     & 0.78  & \textcolor{red}{0.05}  & \textcolor{blue}{0.28} \\ \hline
GPU (GB) & 6.09  & 8.49  & \textcolor{blue}{2.25} \\ \hline
\end{tabular}
\caption{Average testing time and GPU usage of images of size 1280$\times$720 on a single NVIDIA RTX 2080 Ti GPU.}
\label{tab:efficiency_comparison}
\end{table}

The results are shown in Table~\ref{tab:comparison_1}. Our method outperforms most of the existing SOTA methods even without fine-tuning, and fine-tuning on the reblurring network can further improve the performance. In terms of PSNR, Ours+ is ranked first and is 0.07db better than the second~\cite{suin2020spatially}. Although our SSIM is slightly lower than the first~\cite{aljadaany2019douglas}, our PSNR far surpasses it which demonstrates that our method is good at both evaluation metrics. Note that we use the mean squared error loss without a sophisticated GAN~\cite{goodfellow2014generative} and still achieve good performance.

We further compare with some of the methods on the HIDE dataset in Table~\ref{tab:comparison_2}. Ours+ remains the top and Ours is ranked second. Note that unlike all other methods that follow the same strategy of training on the GoPro dataset but being tested on the HIDE dataset, the method of Shen \etal~\cite{shen2019human} is trained on the HIDE dataset directly but it cannot perform better than our proposed method.

The average testing time and GPU memory usage on the GoPro dataset is reported in Table~\ref{tab:efficiency_comparison}. Although the testing time of Park \etal~\cite{park2019multi} is the lowest, its GPU memory usage is almost four times of ours, and its performance (Table~\ref{tab:comparison_1}) is lower than ours. Our method is 30$\%$ faster than Zhang \etal~\cite{zhang2019deep} while only increasing the GPU memory usage by 7$\%$. Since the architecture of Kupyn \etal~\cite{kupyn2019deblurgan} is much deeper and wider than that of other listed methods, its testing time is the highest even if it uses only a single stage. Our proposed method is a good trade-off between performance and efficiency in both memory and computation.

\subsection{Qualitative Comparison}
Following a similar strategy as in the quantitative comparison, we first compare our method against some others on the testing images of the GoPro dataset~\cite{nah2017deep}. We compare with the two best performing and most recent methods~\cite{park2019multi,zhang2019deep} with published well-trained models, along with the published results of Purohit \etal~\cite{purohit2020region} on the same dataset. As apparent in Figure~\ref{fig:qual_gorpo_main}, the output of our method is most similar to the ground truth sharp image, comparing to others. The alphabets written on the banner of the first row are almost clearly visible for ours, while others fail to deblur them correctly. Similarly, the sign over the shop is best deblurred by ours, while the result of Purohit \etal~\cite{purohit2020region} is also reasonably good. On the third row, our method restores the skin and bright colors better. We also use the well-trained models on the GoPro dataset to test on the HIDE dataset~\cite{shen2019human}. As visible in Figure~\ref{fig:qual_hide_main}, our method does a good job deblurring the bicycles in the image. We further test the same models on the RWBI dataset~\cite{zhang2020deblurring}. As seen in Figure~\ref{fig:qual_rwbi_main}, our method generally outperforms the compared methods in terms of deblurring quality. For a more complete qualitative study, please refer to the supplementary material.

\begin{figure}[t]
	\centering
	\begin{subfigure}[h]{0.15\textwidth}
		\includegraphics[width=\textwidth]{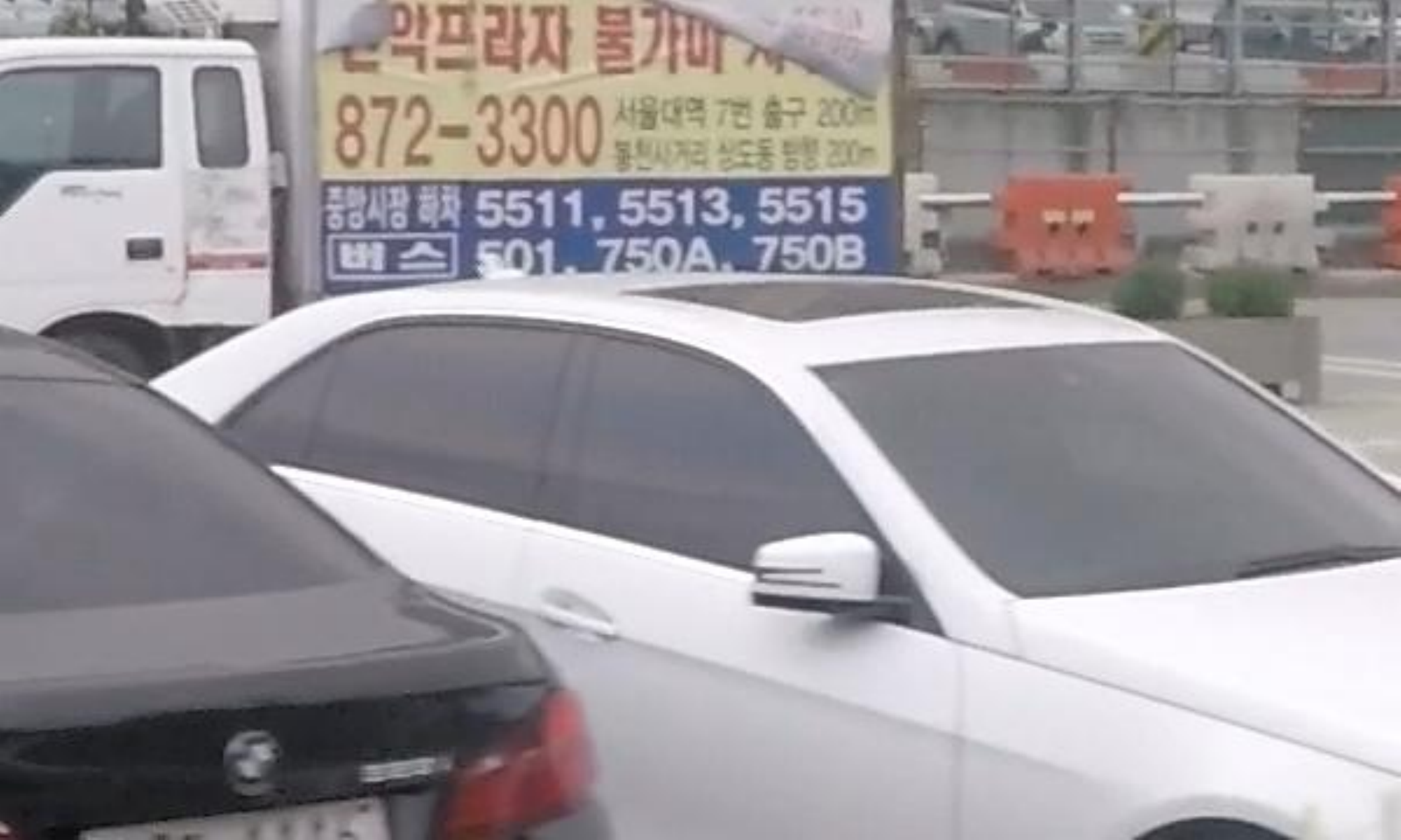}
		\caption{Sharp}
	\end{subfigure}
	\begin{subfigure}[h]{0.15\textwidth}
		\includegraphics[width=\textwidth]{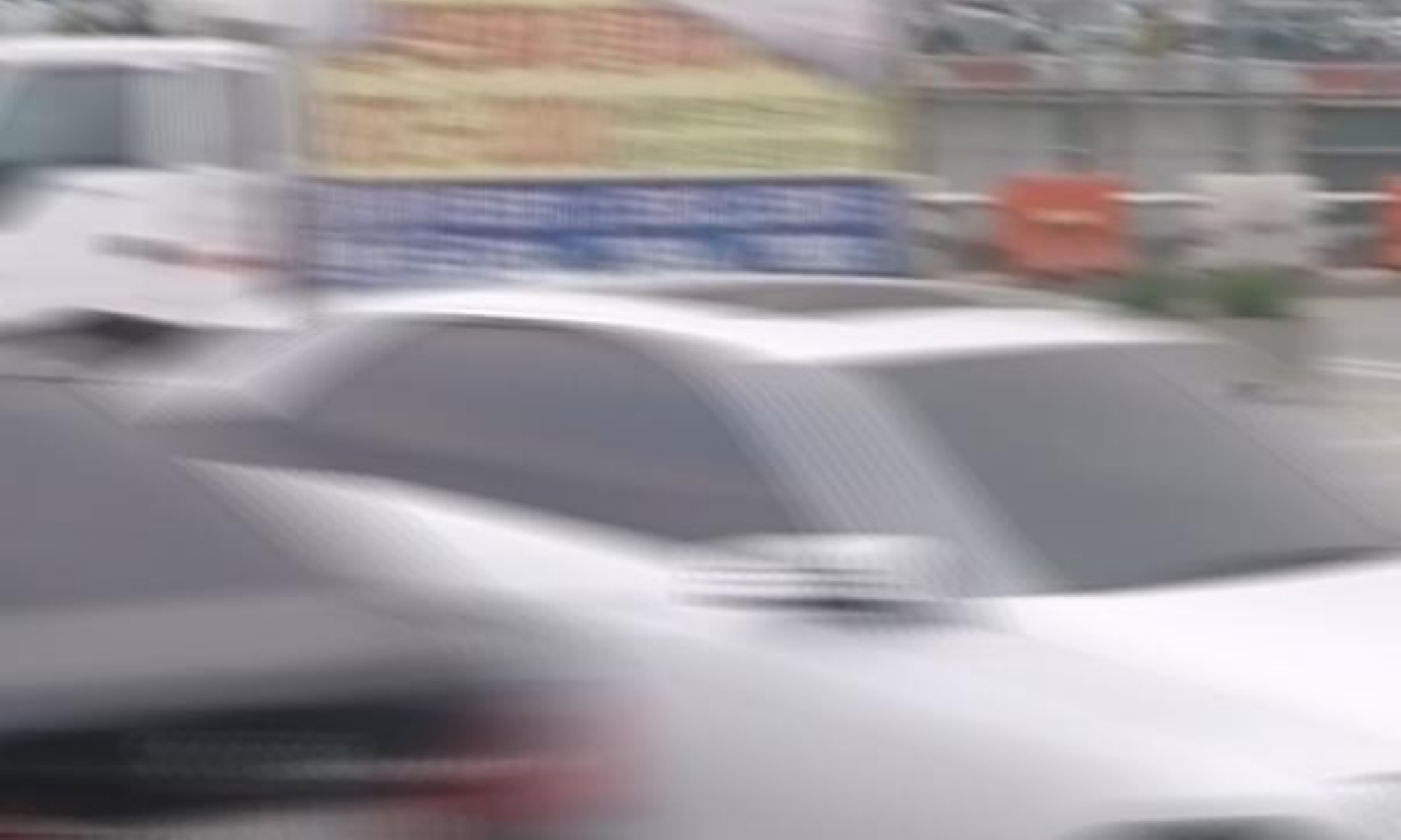}
		\caption{Reblurred}
	\end{subfigure}
	\begin{subfigure}[h]{0.15\textwidth}
		\includegraphics[width=\textwidth]{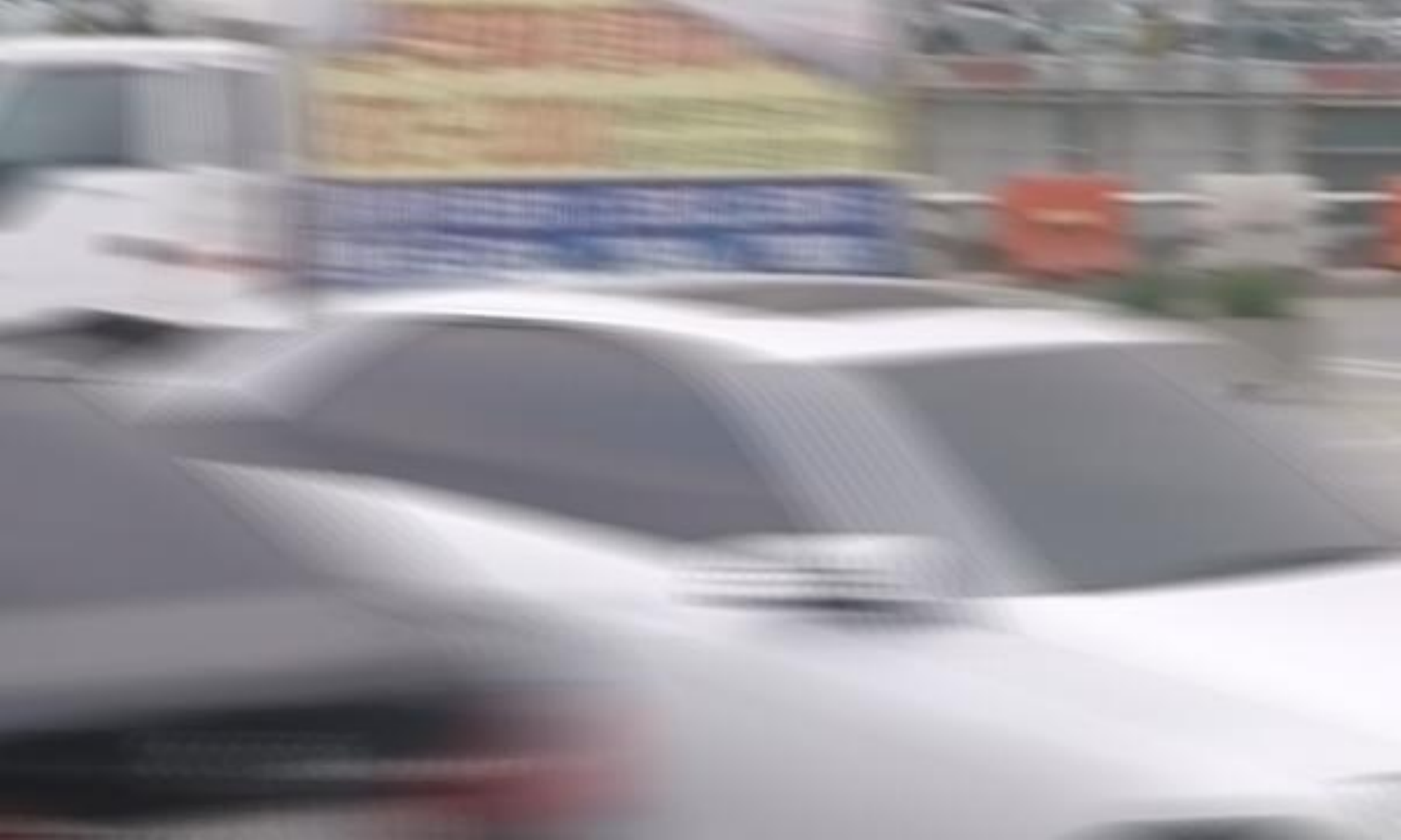}
		\caption{Blurred}
	\end{subfigure}
	
	\begin{subfigure}[h]{0.15\textwidth}
		\includegraphics[width=\textwidth]{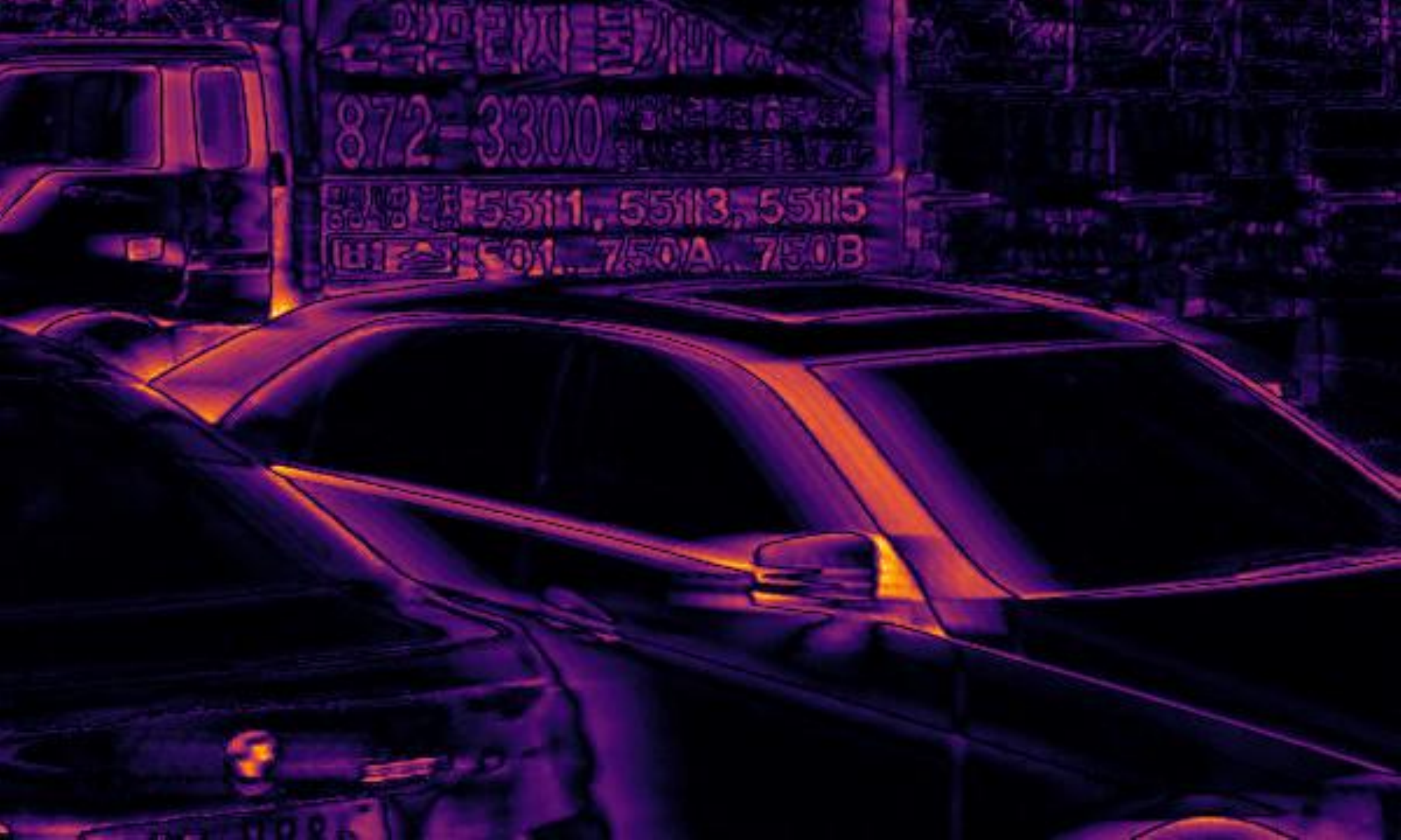}
		\caption{Diff-S}
	\end{subfigure}
	\begin{subfigure}[h]{0.15\textwidth}
		\includegraphics[width=\textwidth]{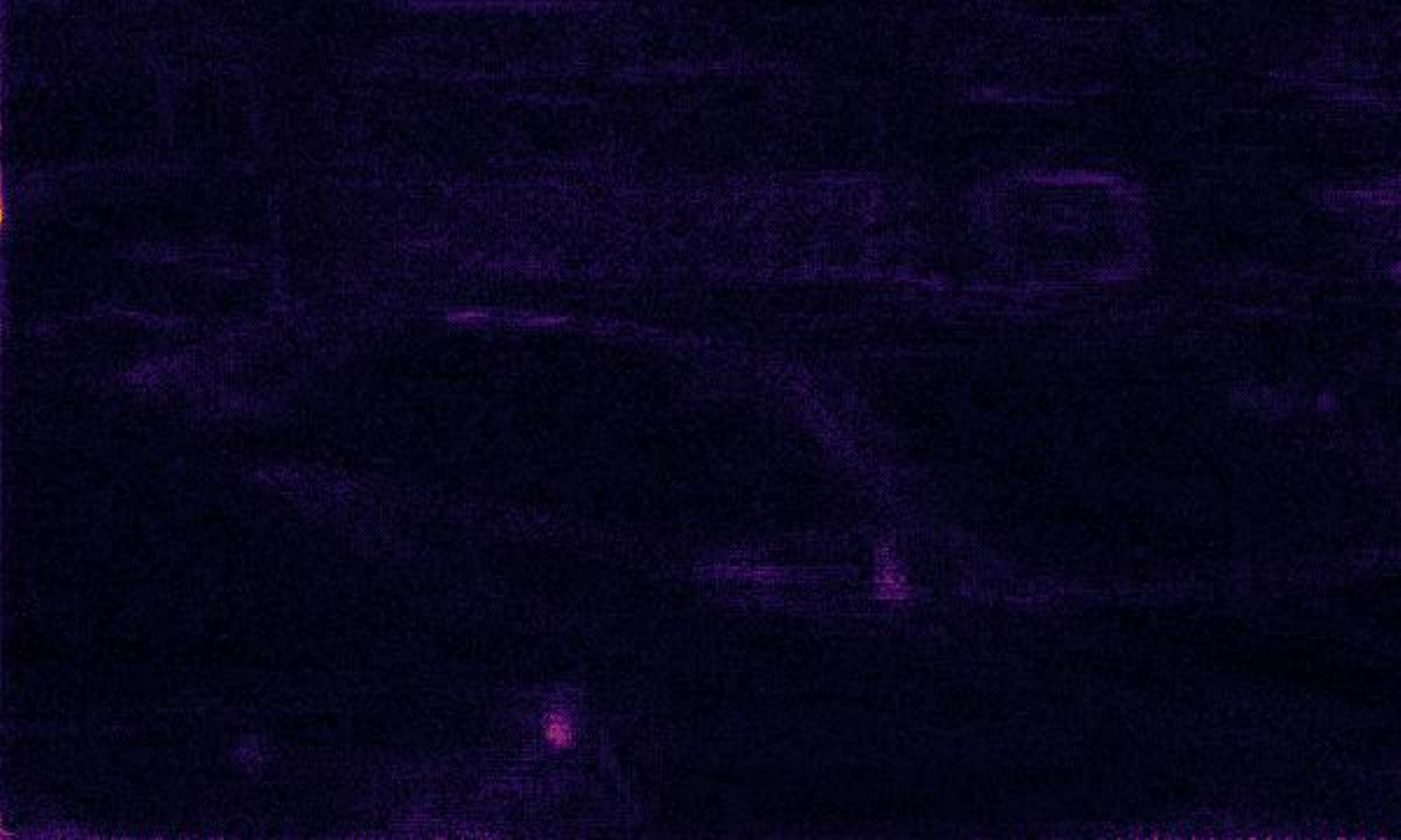}
		\caption{Diff-R}
	\end{subfigure}
	\caption{Comparison of a reblurred output with its corresponding blurred and sharp images. Diff-S (Diff-R) represents the difference map between the sharp image (reblurred output) and the blurred ground truth.}
	\label{fig:Reblurred_result}
\end{figure}

\begin{table}[t]
\centering
\begin{tabular}{|c|c|c|c|}
\hline
PSNR  & SSIM   & Mean   & variance \\ \hline
55.71 & 0.9997 & 0.18 & 0.22            \\ \hline
\end{tabular}
\caption{Performance evaluation of reblurring network on the GoPro dataset.}
\label{tab:reblurring_evaluation}
\end{table}

\begin{figure*}[!htbp]
	\centering
	
	\begin{subfigure}[h]{0.135\textwidth}
		\includegraphics[width=\textwidth]{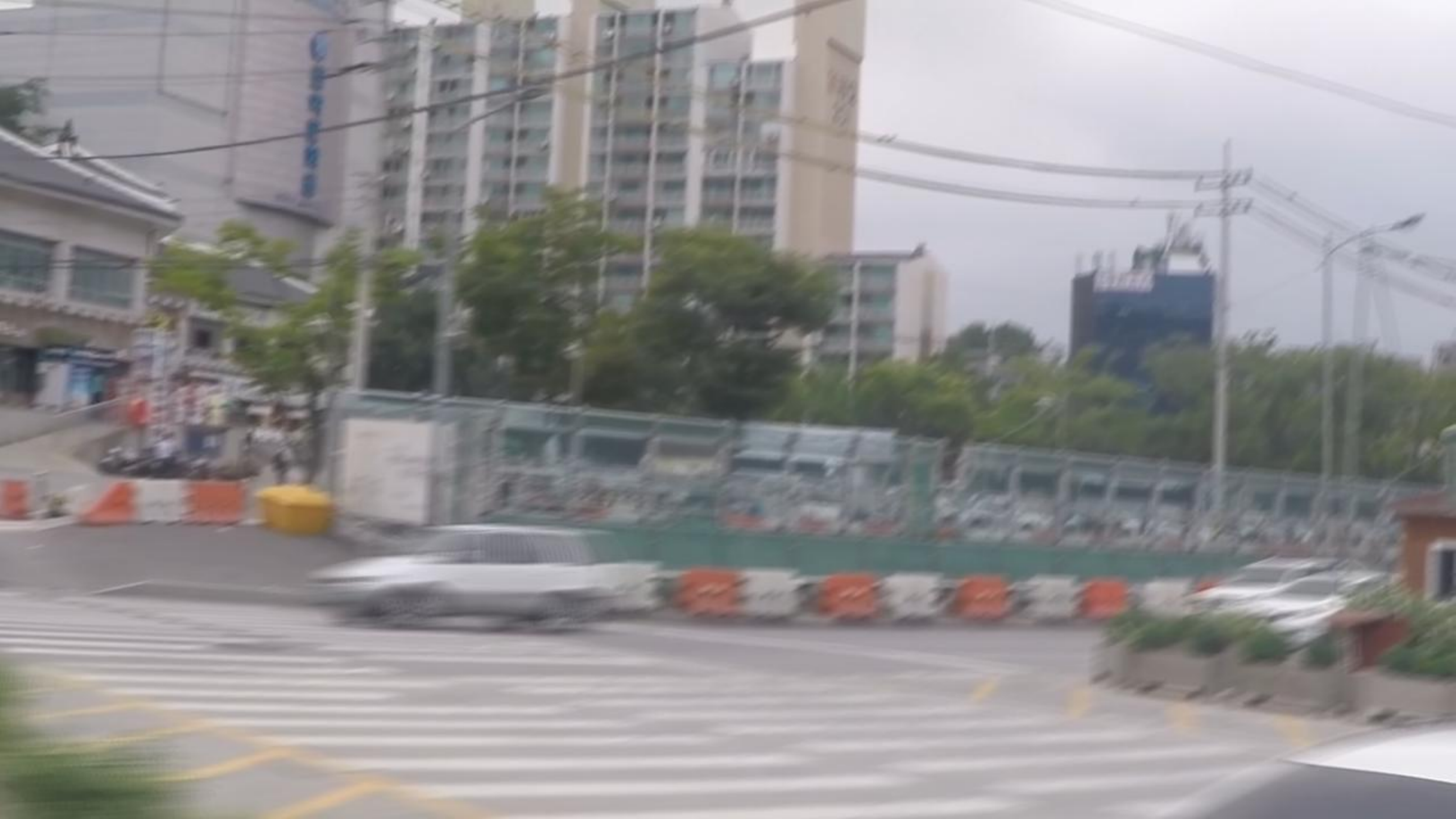}
	\end{subfigure}
	\begin{subfigure}[h]{0.135\textwidth}
		\includegraphics[width=\textwidth]{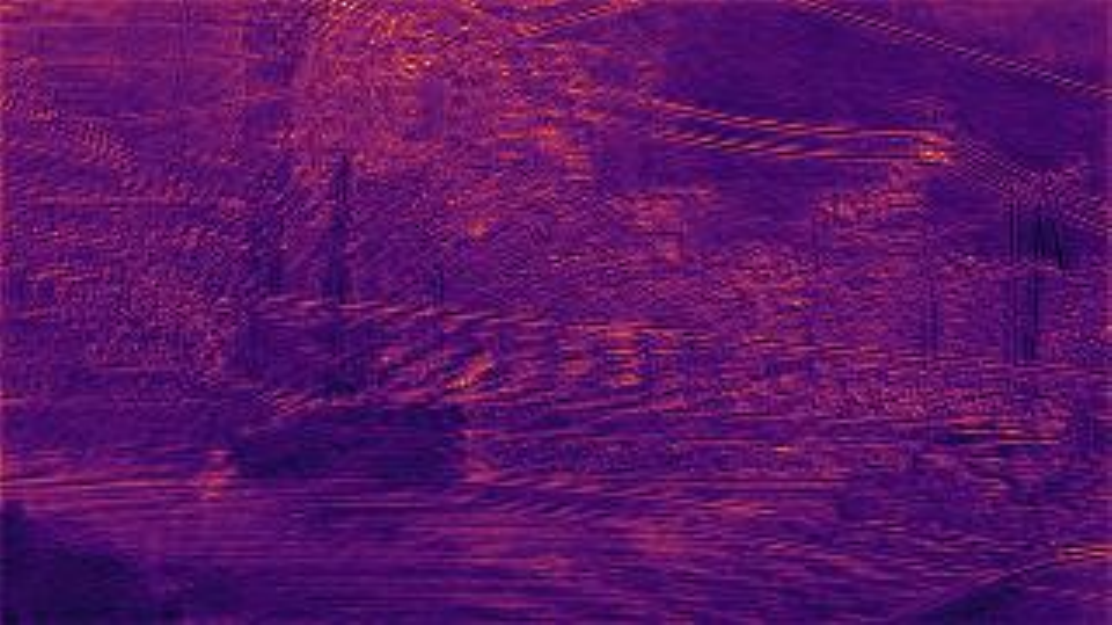}
	\end{subfigure}
	\begin{subfigure}[h]{0.135\textwidth}
		\includegraphics[width=\textwidth]{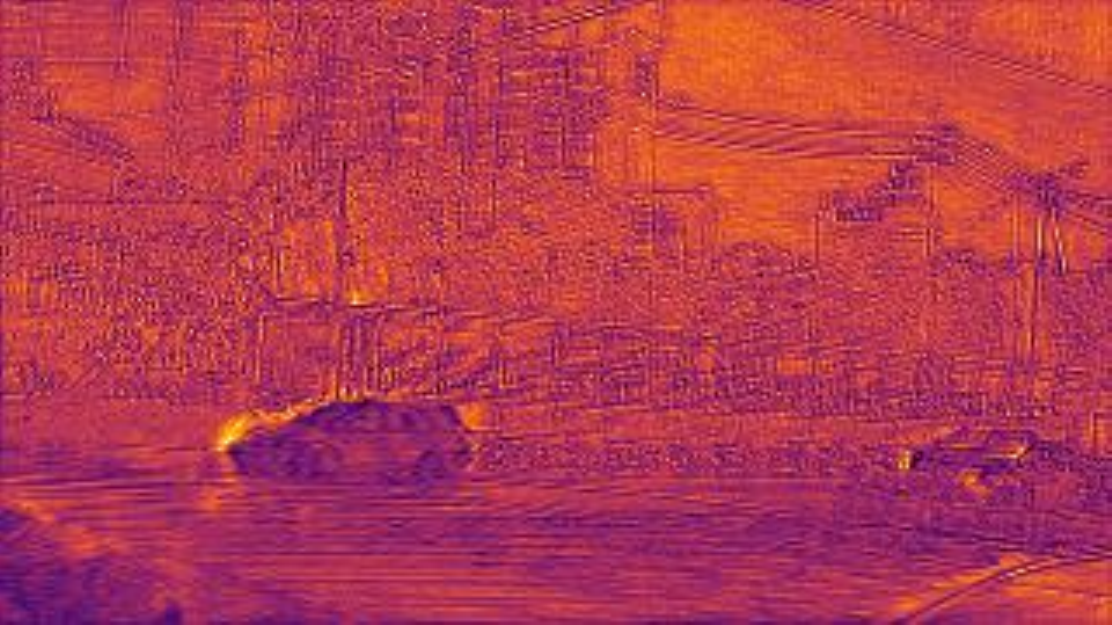}
	\end{subfigure}
	\begin{subfigure}[h]{0.135\textwidth}
		\includegraphics[width=\textwidth]{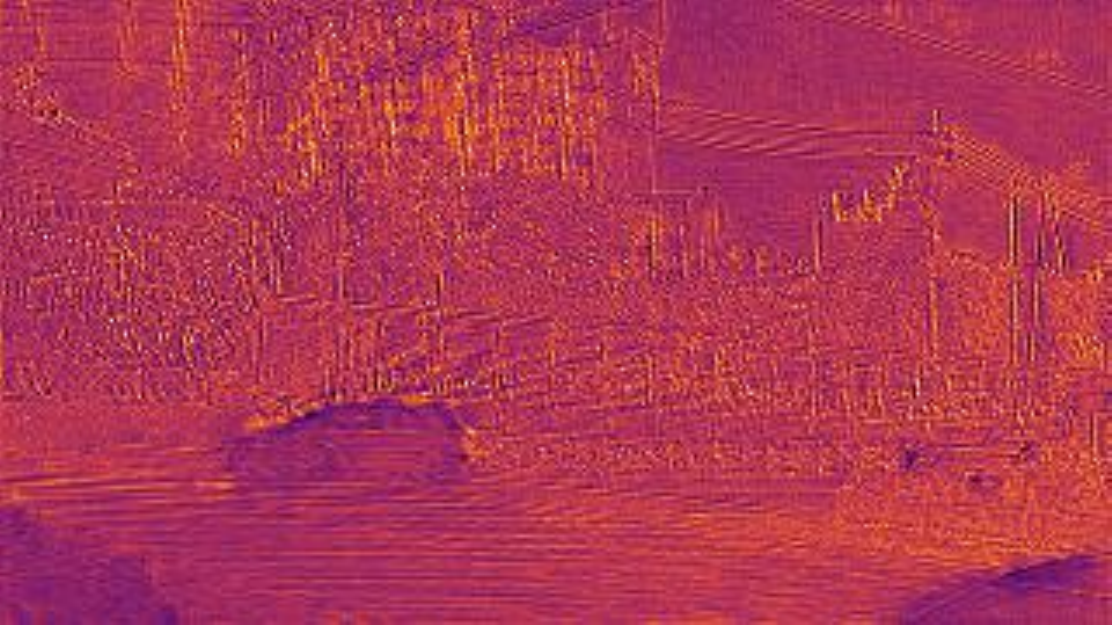}
	\end{subfigure}
	\begin{subfigure}[h]{0.135\textwidth}
		\includegraphics[width=\textwidth]{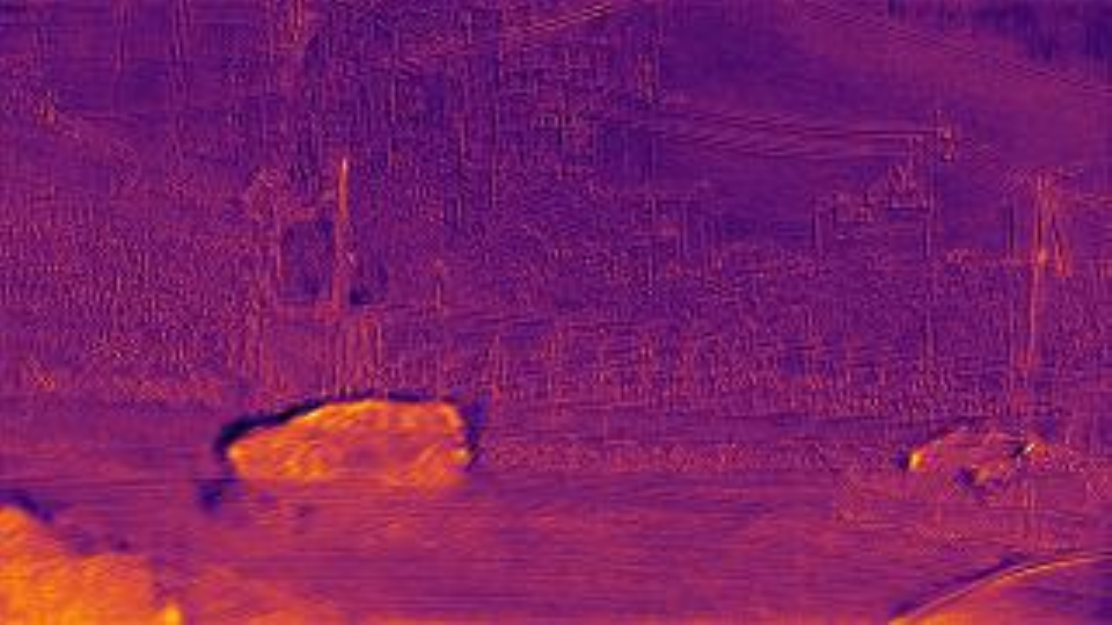}
	\end{subfigure}
	\begin{subfigure}[h]{0.135\textwidth}
		\includegraphics[width=\textwidth]{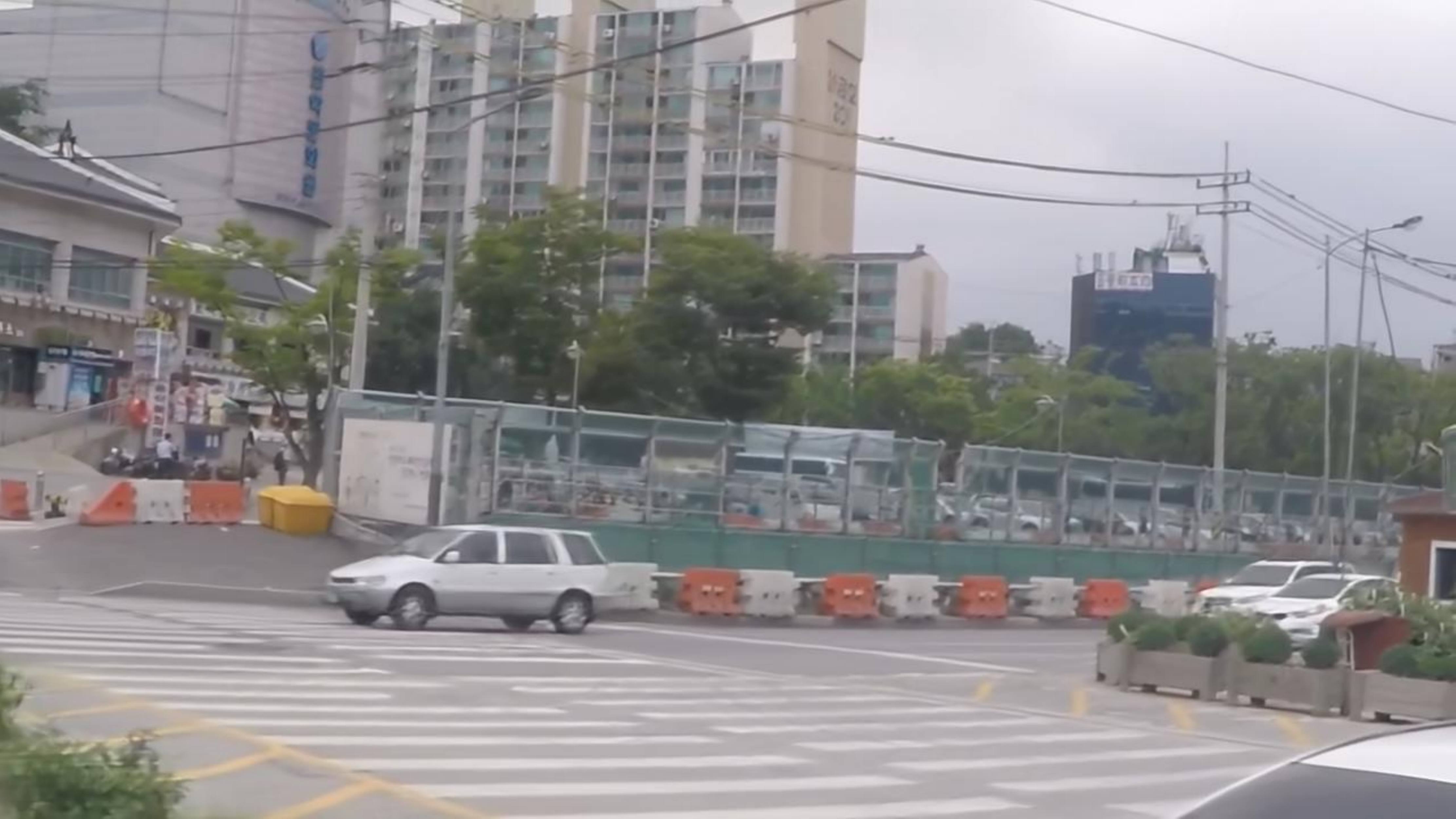}
	\end{subfigure}
	\begin{subfigure}[h]{0.135\textwidth}
		\includegraphics[width=\textwidth]{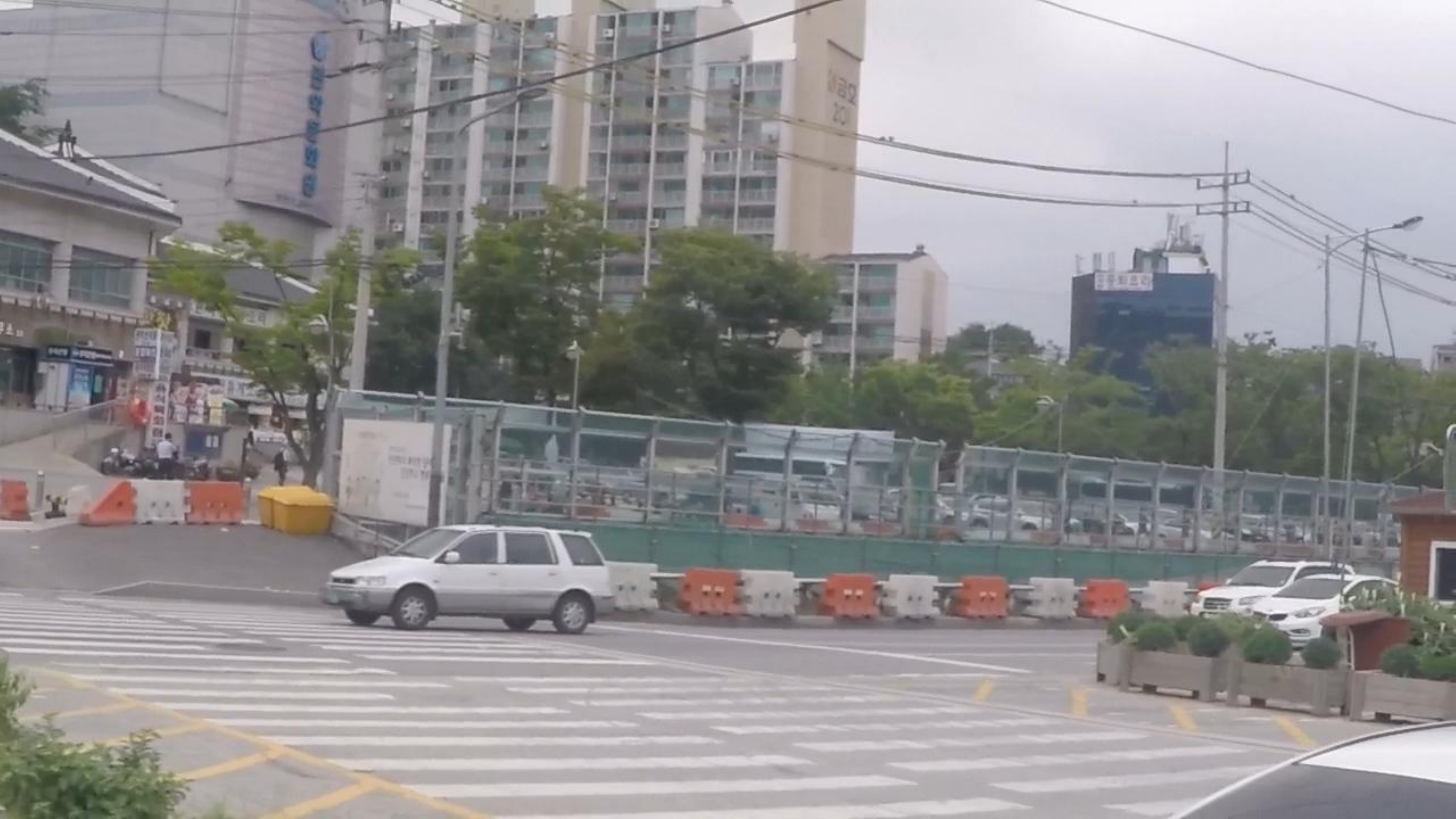}
	\end{subfigure}
	
	\begin{subfigure}[h]{0.135\textwidth}
		\includegraphics[width=\textwidth]{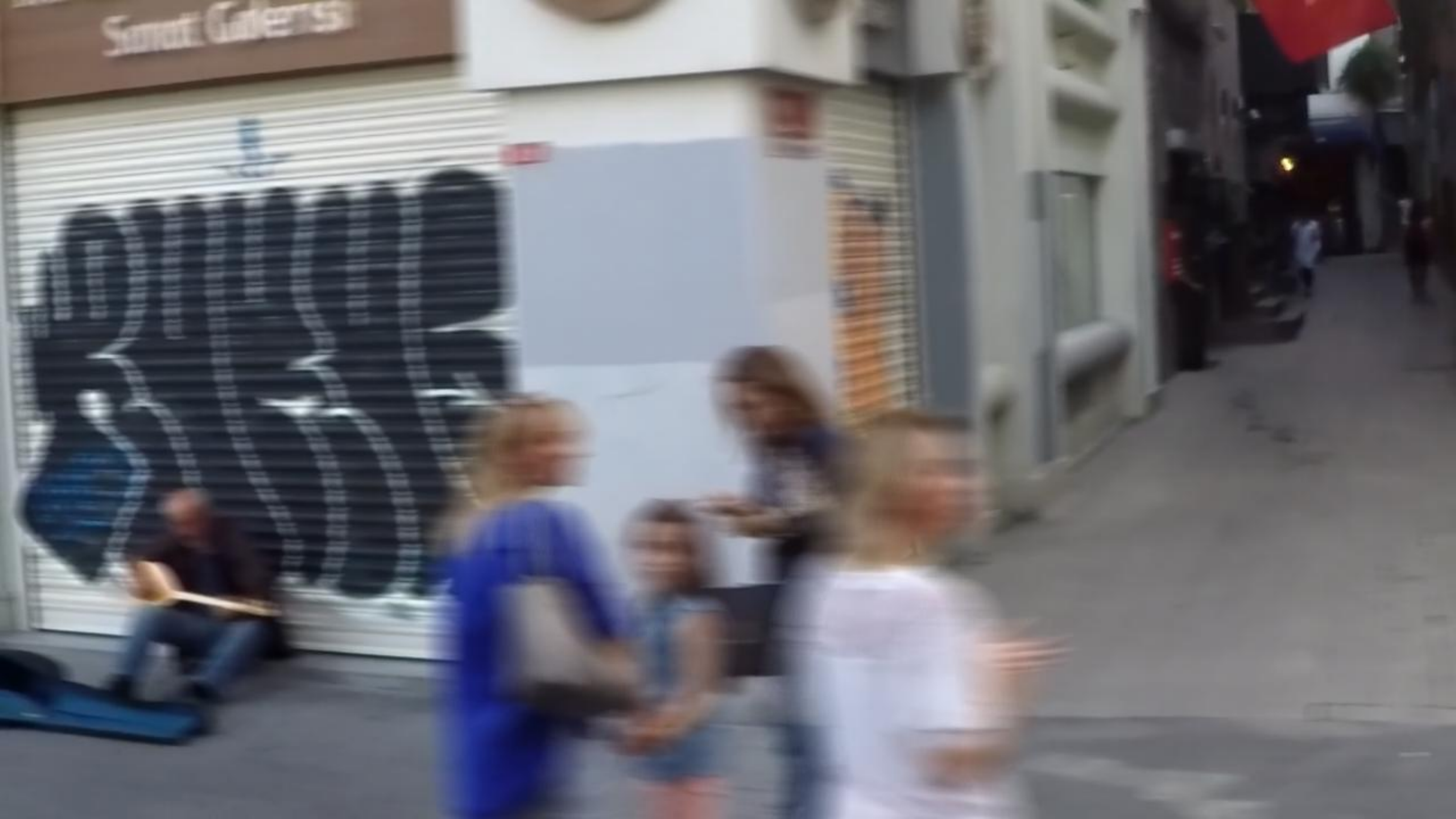}
	\end{subfigure}
	\begin{subfigure}[h]{0.135\textwidth}
		\includegraphics[width=\textwidth]{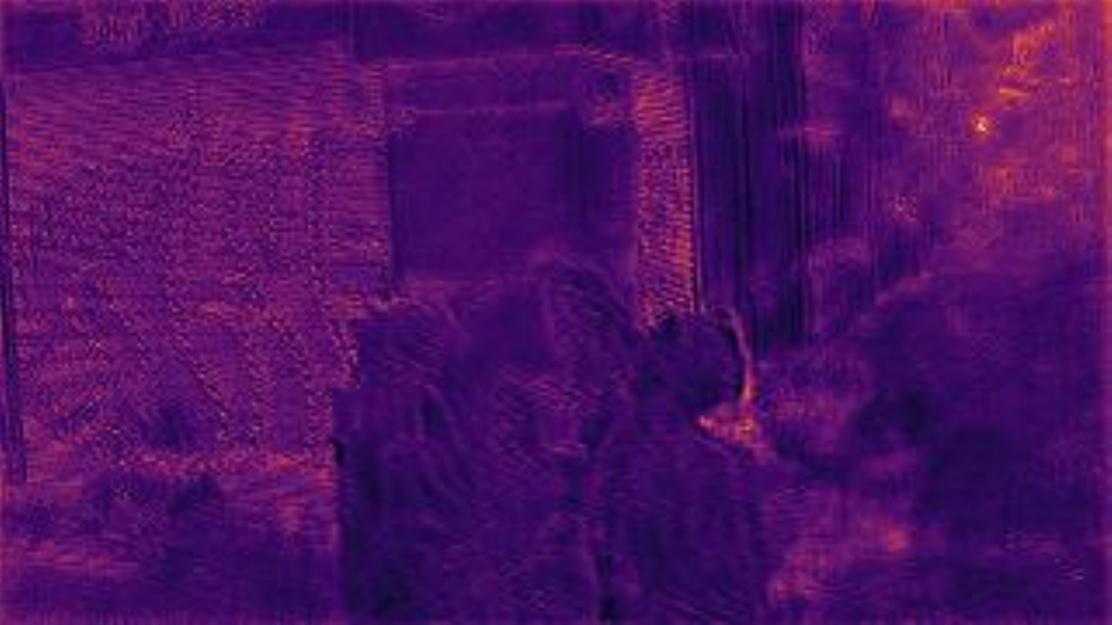}
	\end{subfigure}
	\begin{subfigure}[h]{0.135\textwidth}
		\includegraphics[width=\textwidth]{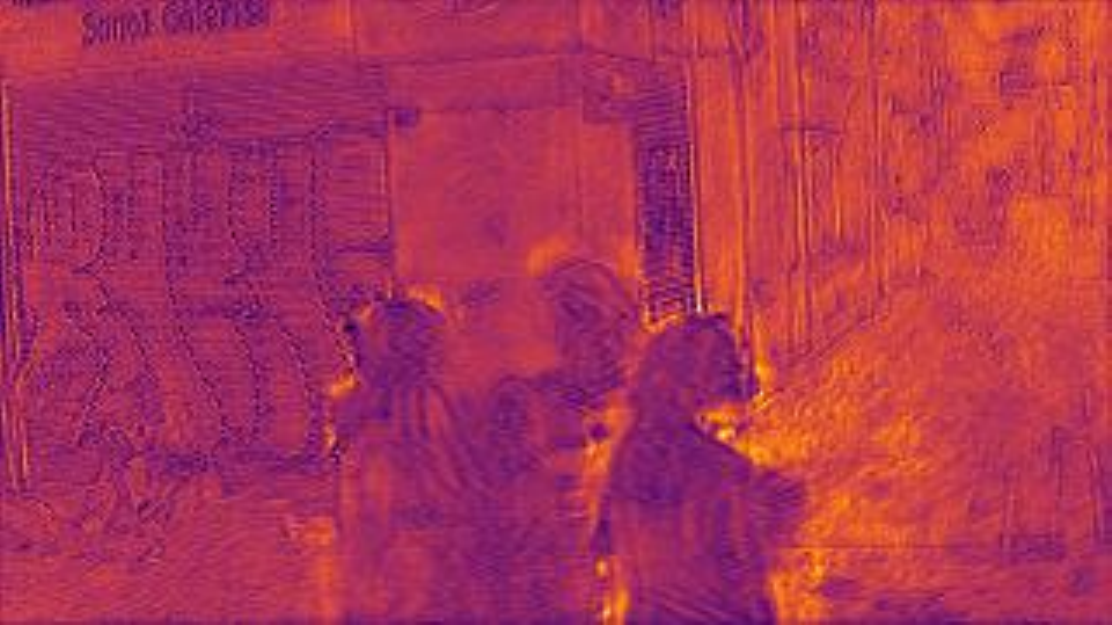}
	\end{subfigure}
	\begin{subfigure}[h]{0.135\textwidth}
		\includegraphics[width=\textwidth]{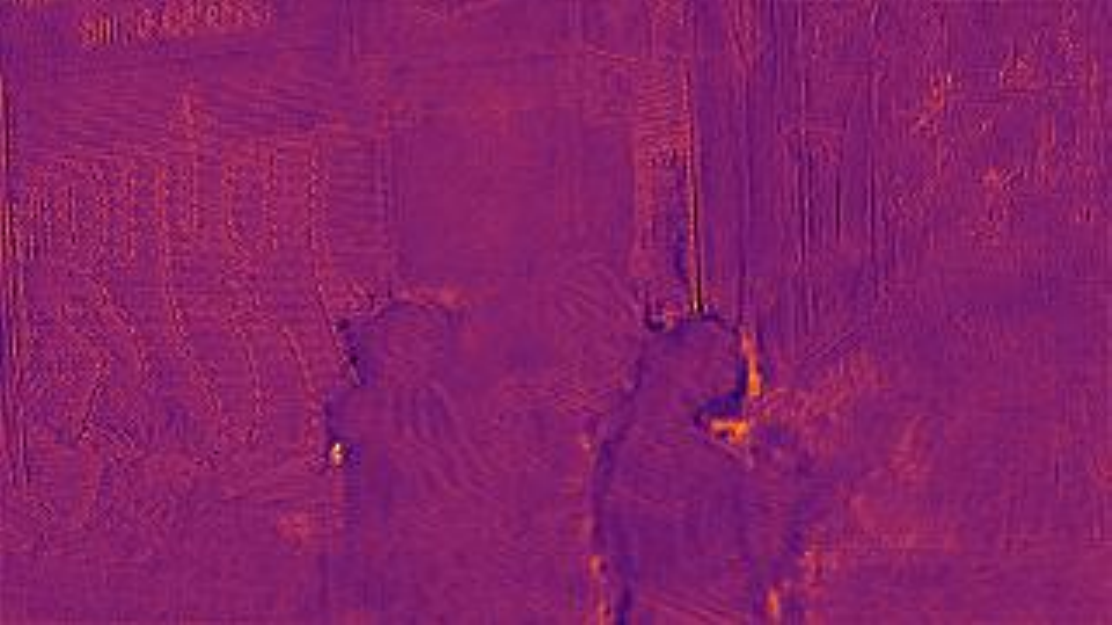}
	\end{subfigure}
	\begin{subfigure}[h]{0.135\textwidth}
		\includegraphics[width=\textwidth]{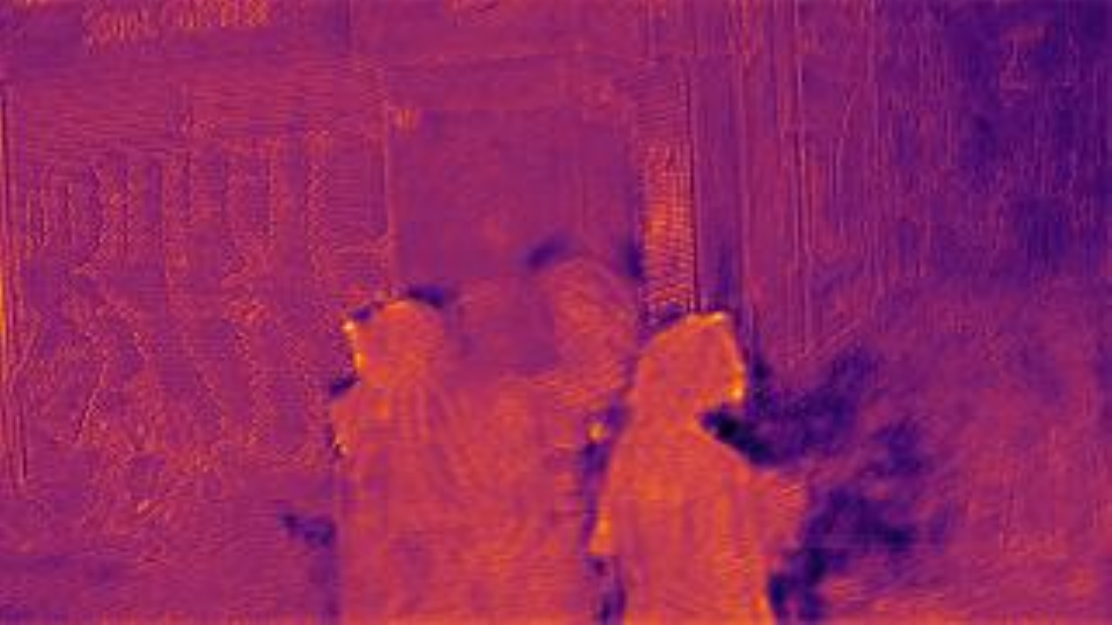}
	\end{subfigure}
	\begin{subfigure}[h]{0.135\textwidth}
		\includegraphics[width=\textwidth]{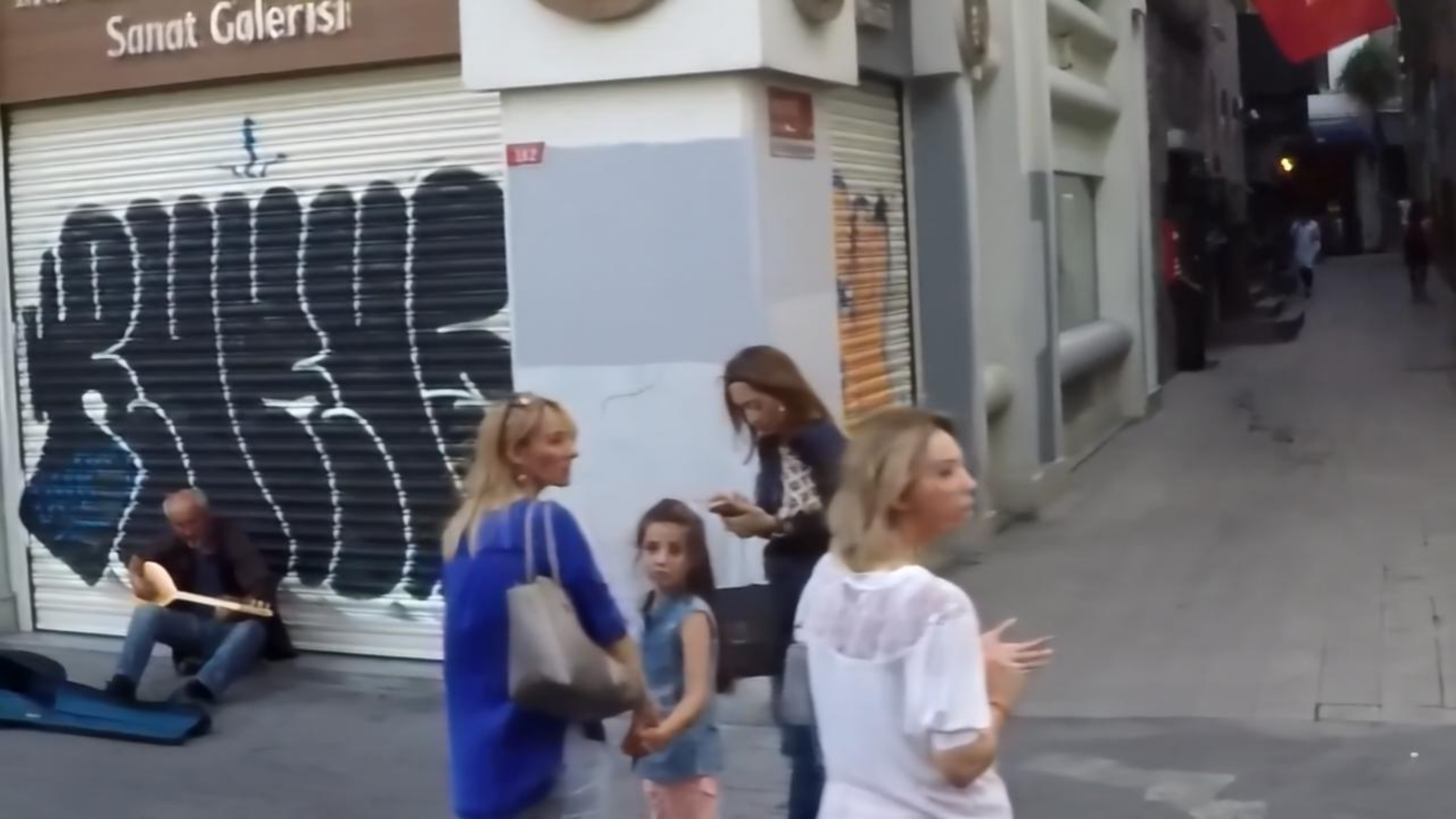}
	\end{subfigure}
	\begin{subfigure}[h]{0.135\textwidth}
		\includegraphics[width=\textwidth]{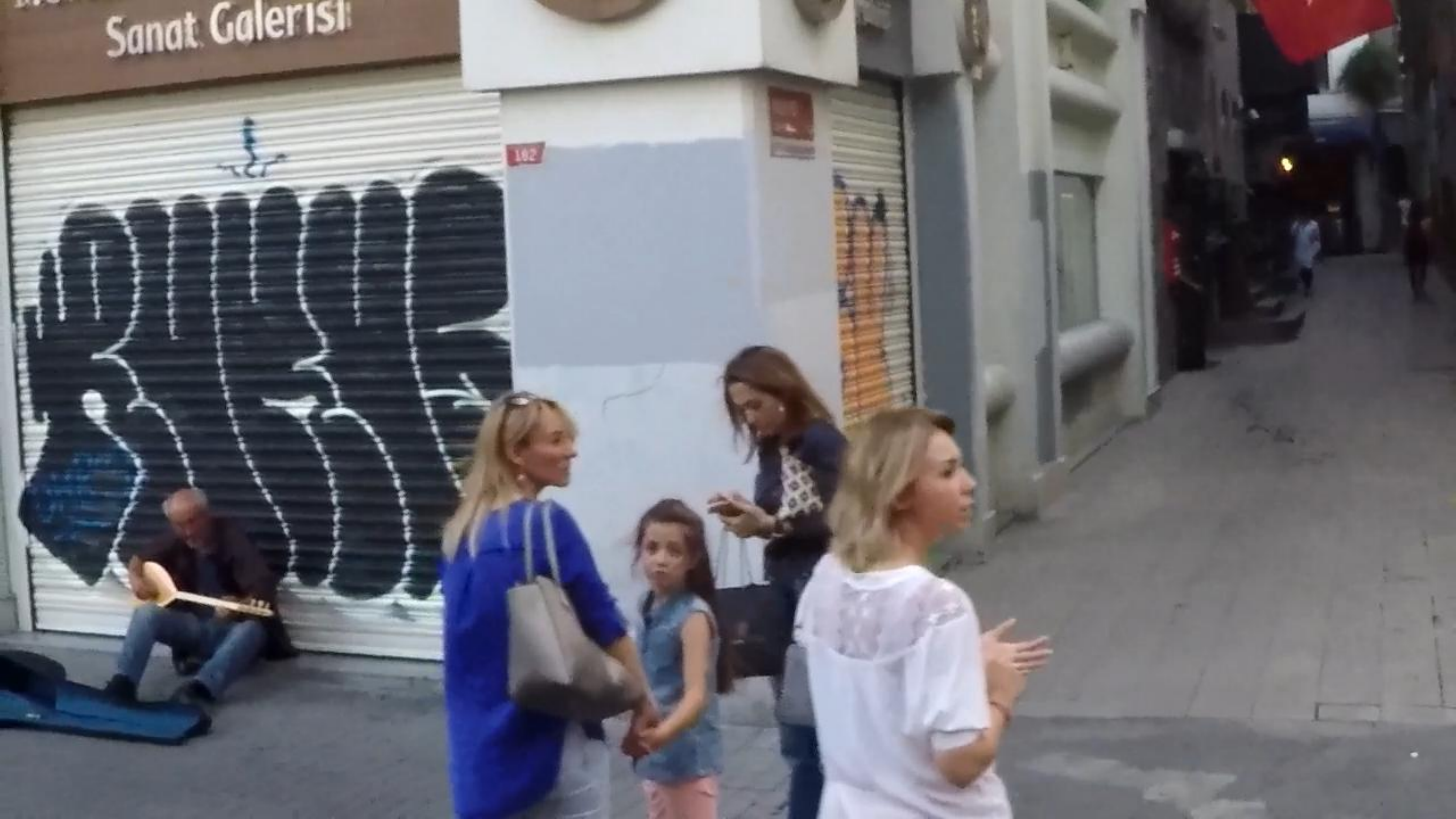}
	\end{subfigure}
	
	\begin{subfigure}[h]{0.135\textwidth}
		\includegraphics[width=\textwidth]{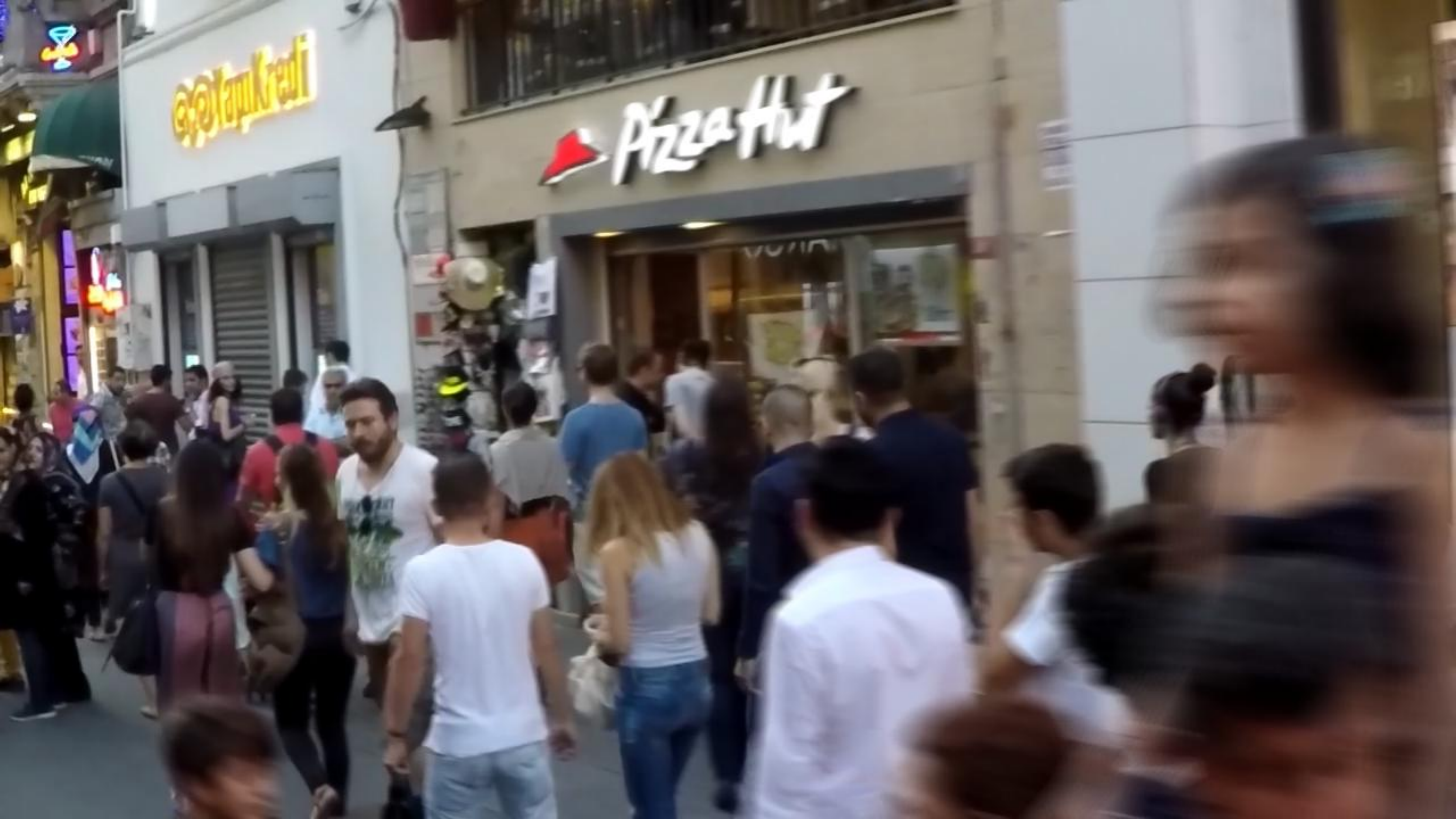}
		\caption{Blurred}
	\end{subfigure}
	\begin{subfigure}[h]{0.135\textwidth}
		\includegraphics[width=\textwidth]{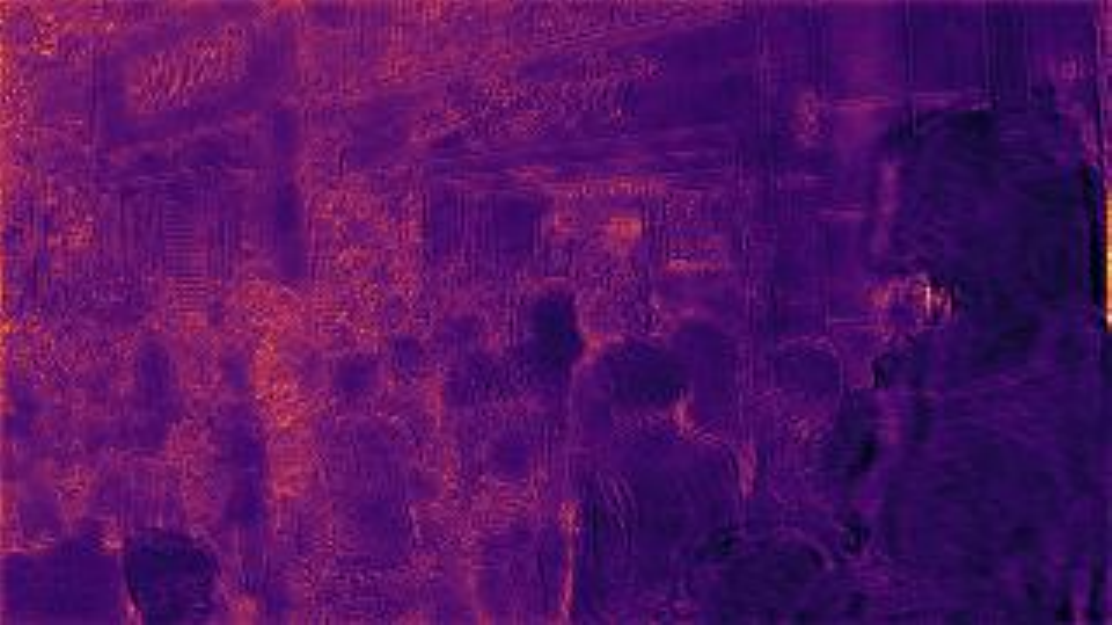}
		\caption{$a_1$}
	\end{subfigure}
	\begin{subfigure}[h]{0.135\textwidth}
		\includegraphics[width=\textwidth]{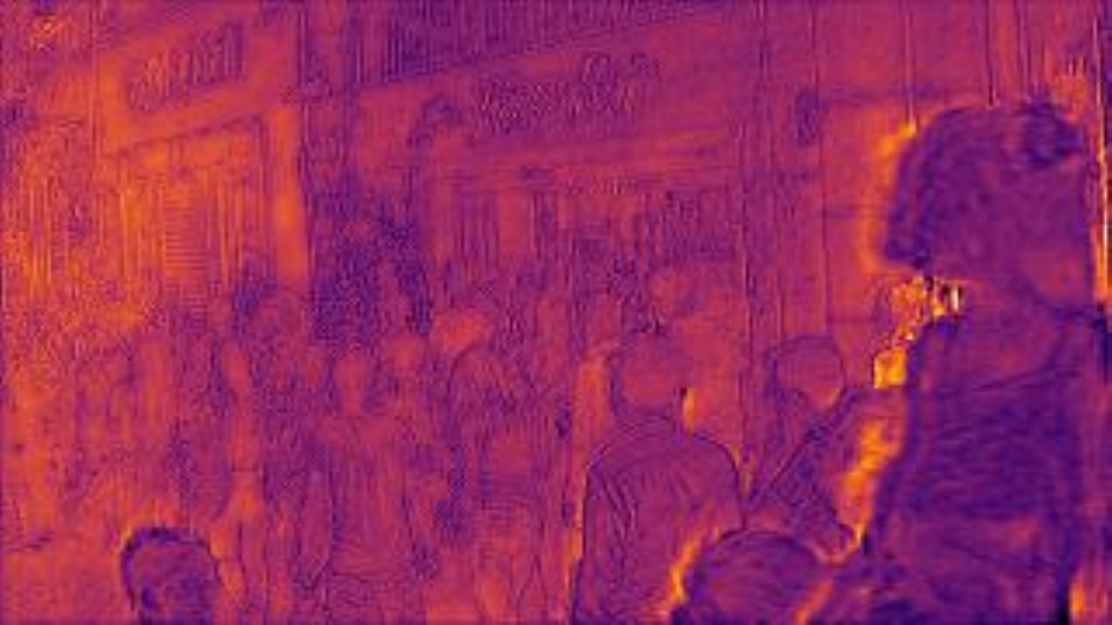}
		\caption{$a_2$}
	\end{subfigure}
	\begin{subfigure}[h]{0.135\textwidth}
		\includegraphics[width=\textwidth]{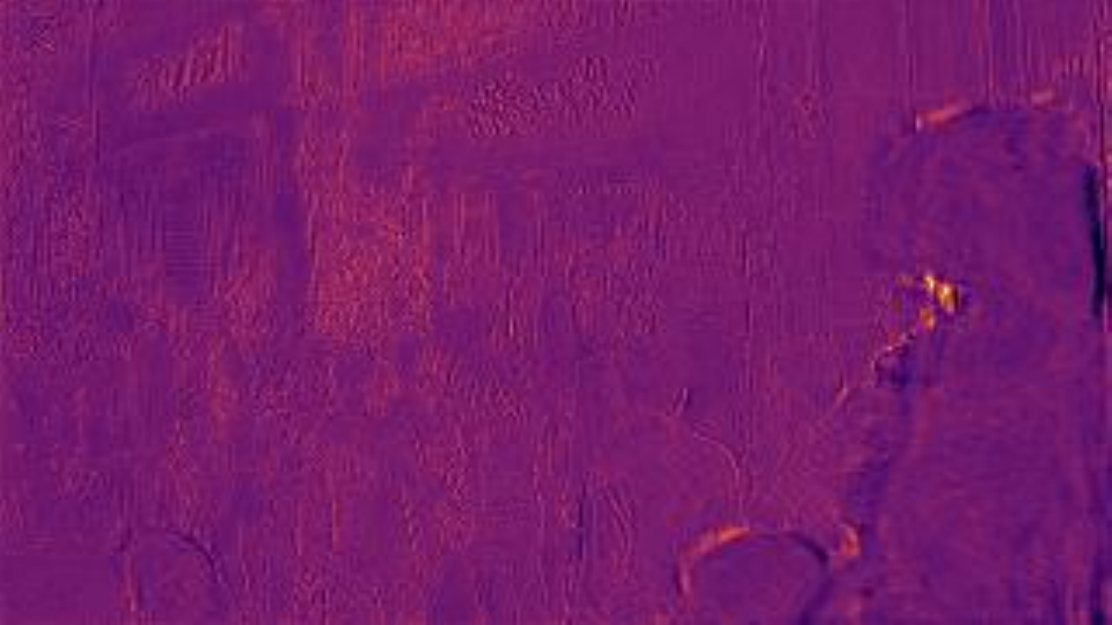}
		\caption{$a_3$}
	\end{subfigure}
	\begin{subfigure}[h]{0.135\textwidth}
		\includegraphics[width=\textwidth]{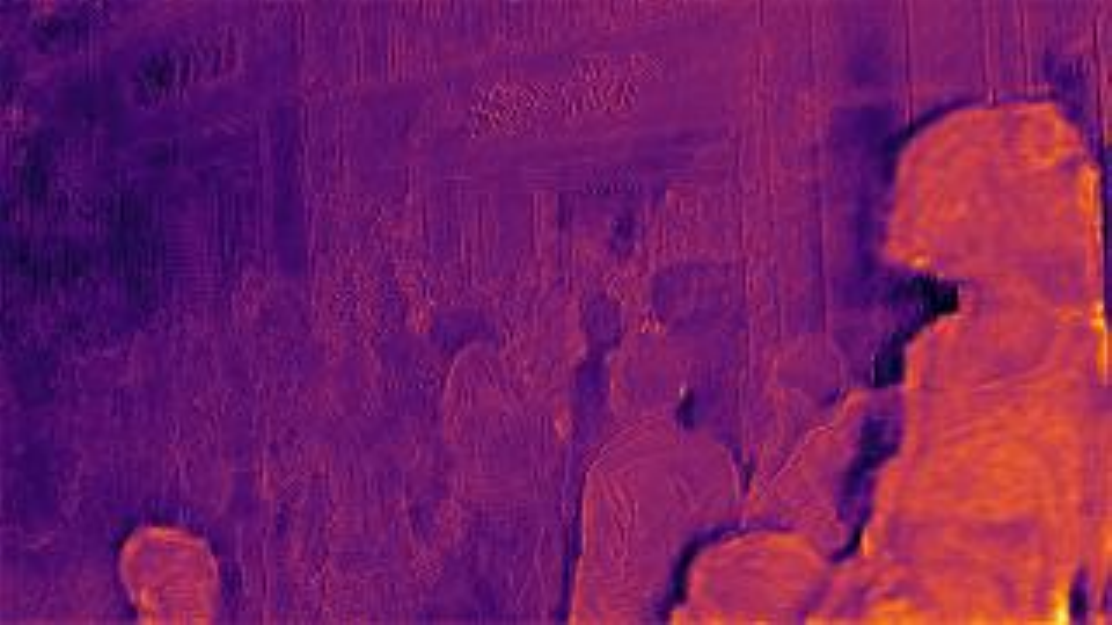}
		\caption{$a_4$}
	\end{subfigure}
	\begin{subfigure}[h]{0.135\textwidth}
		\includegraphics[width=\textwidth]{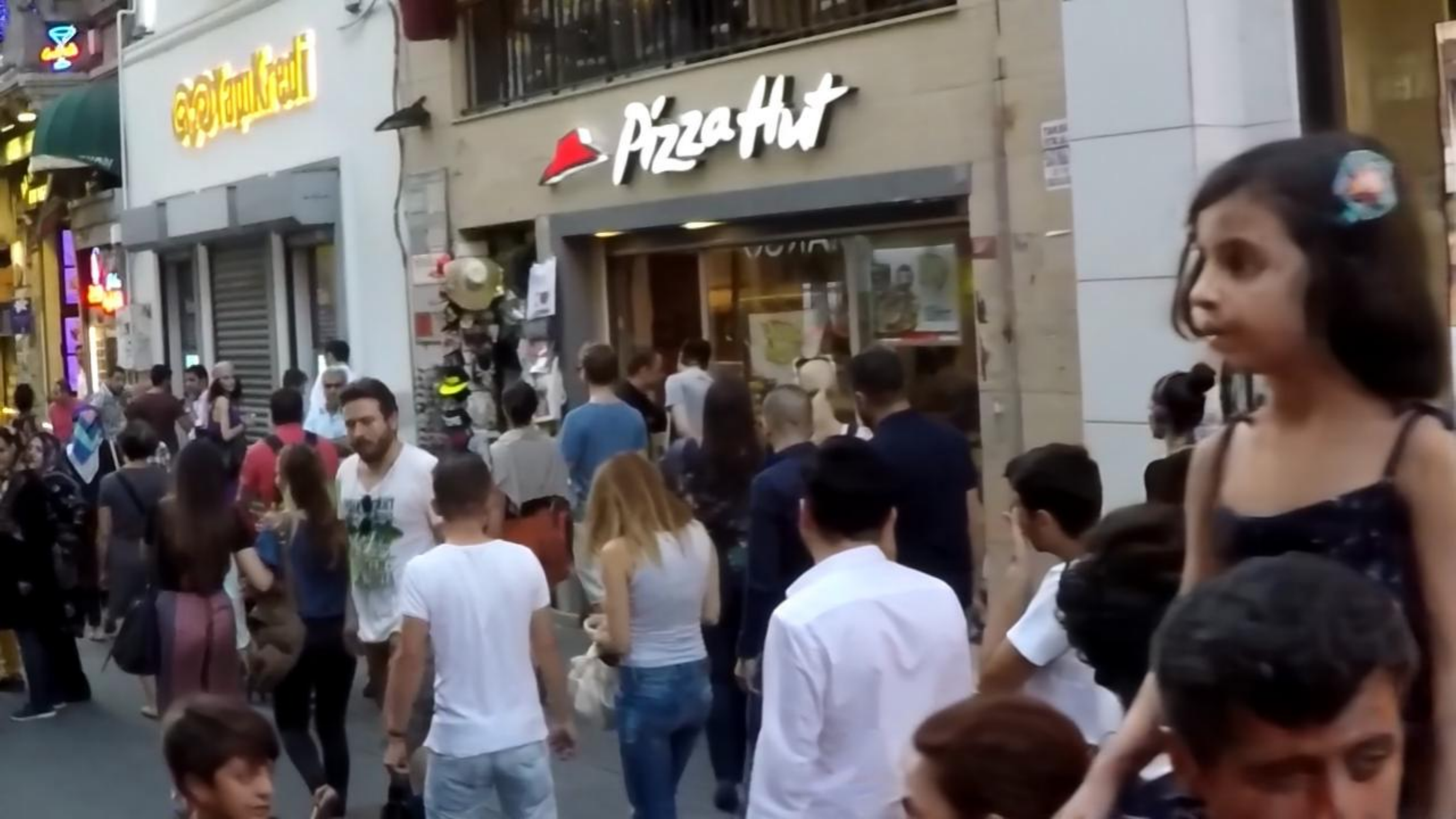}
		\caption{Deblurred}
	\end{subfigure}
	\begin{subfigure}[h]{0.135\textwidth}
		\includegraphics[width=\textwidth]{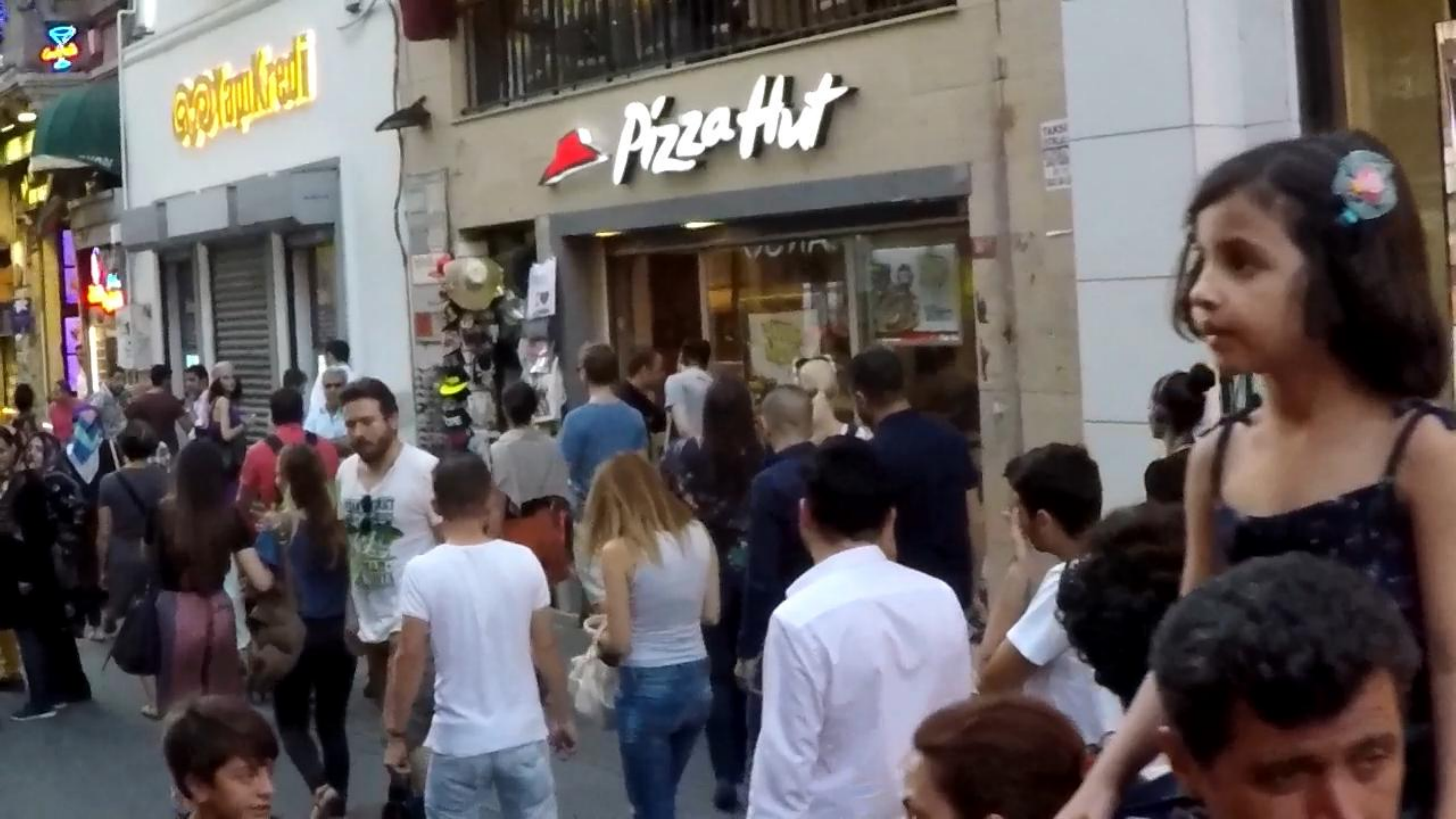}
		\caption{Sharp}
	\end{subfigure}
	\caption{Attention maps of the last ASPDC module. Values are within [0, 1] and the brighter the higher.}
	\label{fig:mask}
\end{figure*}
\begin{table*}[t]
\centering
\begin{tabular}{|c|c|c|c|c|c|c|c|c|c|c|c|c|}
\hline
Version     & (1)       & (2)       & (3)       & (4)       & (5)       & (6)       & (7)       & (8)       & (9)       & (10)      & (11)      & (12)      \\ \hline
Module 1 & \checkmark & \checkmark & \checkmark & \checkmark & \checkmark & \checkmark & \checkmark & \checkmark & \checkmark & \checkmark & \checkmark & \checkmark \\ \hline
Module 2 &           & \checkmark &           &           & \checkmark & \checkmark &           & $\times$3    &           &           & \checkmark & \checkmark \\ \hline
Module 3 &           &           & \checkmark &           & \checkmark &           & \checkmark &           & $\times$3    &           & \checkmark & \checkmark \\ \hline
Module 4 &           &           &           & \checkmark &           & \checkmark & \checkmark &           &           & $\times$3    & \checkmark & \checkmark \\ \hline
AFIM                &           & \checkmark & \checkmark & \checkmark & \checkmark & \checkmark & \checkmark & \checkmark & \checkmark & \checkmark &           & \checkmark \\ \hline\hline
PSNR                & 30.24     & 30.65     & 30.94     & 30.85     & 31.82     & 31.73     & 31.53     & 31.83     & \textcolor{blue}{31.93}     & 31.72     & 31.16     & \textcolor{red}{32.12}     \\ \hline
SSIM                & 0.942     & 0.944     & 0.948     & 0.947     & 0.957     & 0.956     & 0.954     & 0.957     & \textcolor{blue}{0.958}     & 0.955     & 0.950     & \textcolor{red}{0.959}     \\ \hline
\end{tabular}
\caption{Performance of different versions of the ASPDC module.}
\label{tab:ablation_study}
\end{table*}

\begin{table}[t]
\centering
\begin{tabular}{|c|c|c|c|c|}
\hline
$\lambda$ & 0.01 & 0.1   & 1    \\ \hline
PSNR  & \textcolor{blue}{32.17} & \textcolor{red}{32.22} & 32.15 \\ \hline
SSIM  & \textcolor{red}{0.960} & \textcolor{red}{0.960} & \textcolor{blue}{0.959}   \\ \hline
\end{tabular}
\caption{Performance of fine-tuning with different $\lambda$.}
\label{tab:ablation_study2}
\end{table}

\subsection{Reblurring Evaluation}
In addition to evaluation of the deblurring network, we evaluate the performance of the reblurring network since it is critical for the fine-tuning. As shown in Table~\ref{tab:reblurring_evaluation}, PSNR and SSIM of reblurred images are quite high, and the mean and variance of the difference map $| I_r - I_b |$ are almost 0. The experimental result illustrates that reblurred images are quite close to the original blurred images. 

One of the reblurred outputs is shown in Figure~\ref{fig:Reblurred_result} with the original sharp and blurred images. As we can see, the difference between the reblurred and blurred image is small enough and negligible.

\subsection{Ablation Study}
To evaluate the effectiveness of each component in the ASPDC module, we compare multiple versions of it. We randomly select 200 testing images from the GoPro dataset for the validation. All versions are trained with the mean squared error loss only without fine-tuning. With regard to the performance and efficiency trade-off, we find having four ASPDC modules is optimal.

The experimental results are shown in Table~\ref{tab:ablation_study}. Version 1 has a deformable module 1 without an offset $\Delta p$, which is used as our baseline. The $\times$3 in Version 8$\sim$10 represents three duplicated modules (the same dilation rate but no parameter-sharing). As we can see, the deformable module 2$\sim$4 is critical for improving performance, especially module 3. Simply duplicating modules cannot get results as good as combining different modules. It demonstrates that fusing information from different sizes of receptive fields is meaningful and justified. Version 11 demonstrates that features from different receptive fields should be fused properly. We visualize the attention maps of the last (sixth) ASPDC module in Figure~\ref{fig:mask}. It shows that attention maps of small receptive fields ($a_1$ and $a_2$) focus more on static objects or objects with small movements, while attention maps of large receptive fields (especially $a_4$) pay more attention on objects with large movements.

For the choice of the hyperparameter $\lambda$ in Eqn~\ref{eqn:overall_loss}, we evaluate values in the range of 0.01 to 1 in Table~\ref{tab:ablation_study2}. As we can see, the deblurring term is overwhelmed by the reblurring term when the value of $\lambda$ is too large. Inversely, a small value of $\lambda$, such as 0.01, limits the effect of the reblurring term on the improvement of performance. To fully utilize the reblurring term, we set $\lambda=0.1$ in all our experiments.

\section{Conclusion}
In this paper, we propose a novel end-to-end blind non-uniform motion deblurring network with new ASPDC modules, which are able to apply region-specific convolution to each pixel and integrate features from different receptive fields. Compared to SOTA methods, our method achieves better performance with high efficiency. In addition, the performance can be further improved by fine-tuning on the proposed reblurring network. In future, we plan to address the fact that none of the existing methods perform well when the magnitude of motion gets too large, resulting in issues such as color degradation or even failure in deblurring. We further plan to study deblurring-reblurring consistency of non-uniform deblurring in an unsupervised setting with no access to blurred-sharp pairs for the reblurring network.

{\small
\bibliographystyle{ieee_fullname}
\bibliography{egbib}
}

\end{document}